\documentclass{article}

\PassOptionsToPackage{numbers,sort&compress}{natbib}
\usepackage[main, final]{neurips_2026}

\usepackage[utf8]{inputenc}
\usepackage[T1]{fontenc}
\usepackage{hyperref}
\hypersetup{
  hidelinks,
  pdftitle={SoccerNarrate: Event-Grounded Streaming Soccer Commentary with Macro-Window Preference Alignment},
  pdfauthor={Zihan Jia, Zhilin Dai, Zhengming Zhang, Min Yang, Zhenpeng Huang, Jiaqi Li, Caixia Sun, Xi Chen, Liang Li, Junlan Feng, Gangshan Wu, Limin Wang}
}
\usepackage{url}
\usepackage{booktabs}
\usepackage{amsfonts}
\usepackage{nicefrac}
\usepackage{microtype}
\usepackage[table]{xcolor}

\usepackage{subcaption}
\usepackage{diagbox}
\usepackage{amssymb}
\usepackage{amsmath}
\usepackage{multirow}
\usepackage{caption}
\usepackage{cuted}
\newcommand{\xmarklight}{\textcolor{gray!55}{\ding{55}}}
\usepackage{tabularx}
\usepackage{mathtools}
\usepackage{amsthm}
\usepackage{pifont}
\newcommand{\cmark}{\ding{51}}

\usepackage{makecell}
\usepackage{graphicx}
\usepackage[capitalize,noabbrev]{cleveref}
\theoremstyle{plain}

\theoremstyle{definition}

\theoremstyle{remark}

\title{{\fontsize{15.8}{20}\selectfont SoccerNarrate: Event-Grounded Streaming Soccer\\Commentary with Macro-Window Preference Alignment}}

\author{%
  Zihan Jia$^{1}$\thanks{These authors contributed equally to this work.}, Zhilin Dai$^{1}$\footnotemark[1], Zhengming Zhang$^{2}$\footnotemark[1], Min Yang$^{1}$, Zhenpeng Huang$^{1}$, Jiaqi Li$^{2}$,\\
  \textbf{Caixia Sun$^{2}$, Xi Chen$^{2}$, Liang Li$^{2}$, Junlan Feng$^{2}$, Gangshan Wu$^{1}$, Limin Wang$^{1}$\thanks{Corresponding author (\texttt{lmwang@nju.edu.cn}).}}\\
  $^{1}$State Key Laboratory for Novel Software Technology, Nanjing University\\
  $^{2}$China Mobile Jiutian Artificial Intelligence Technology (Beijing) Co., Ltd.
}

\begin{document}

\maketitle

\vspace{-1.5em}
\noindent\begin{minipage}{\textwidth}
\centering
\includegraphics[width=0.98\textwidth]{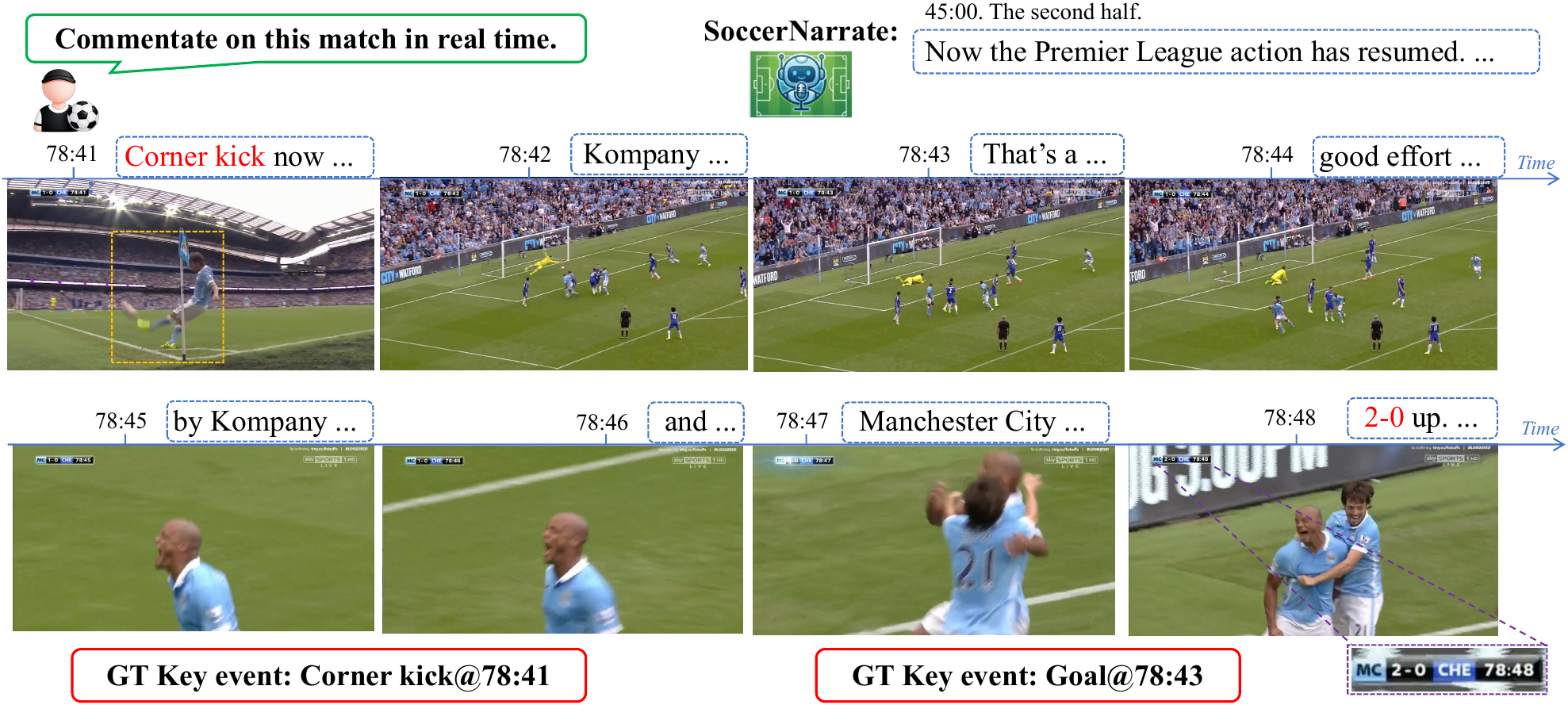}

\parbox{0.98\textwidth}{\footnotesize
Figure 1. SoccerNarrate generates real-time, event-grounded soccer commentary for streaming broadcasts, emulating a human commentator.
Given a video stream, it produces continuous commentary with a latency less than 0.3s per response on a single NVIDIA H100, supporting real-time applications at 2 FPS.}
\end{minipage}

\begin{abstract}
Existing clip-based soccer commentary models often produce delayed, repeated, or poorly synchronized narration on untrimmed full-match videos. Recent streaming video-language models make low-latency full-match narration feasible, but they often favor general descriptions rather than professional, event-grounded soccer commentary. They may describe nearby actions while failing to mention key events such as goals, cards, substitutions, and offsides correctly and on time. In this paper, we present \textbf{SoccerNarrate} to bridge low-latency streaming narration with event-grounded professional soccer commentary across data, training, and evaluation. First, we construct \textbf{SoccerNarrate-Data}, a large-scale time-anchored dataset with word-level ASR alignment, quality filtering, and entity calibration. Second, we perform \textbf{macro-window preference alignment (MWPA)}, which aligns the model with complete event-level commentary semantics by comparing causal multi-step rollouts from the same streaming prefix while keeping second-level inference unchanged. We further use event-mismatched human commentaries as counterfactual negatives to emphasize event semantics over commentary style. Third, we introduce \textbf{SoccerNarrate-Eval}, an event-centric benchmark based on temporally constrained event entailment. Full-match experiments show that SoccerNarrate improves event coverage and event-aligned precision over strong offline and streaming baselines. Code and models are released at \url{https://github.com/MCG-NJU/SoccerNarrate}.
\end{abstract}

\section{Introduction}

Building on broader progress in sports video understanding---such as spatio-temporal action detection, player tracking, and human-human interaction modeling~\cite{li2021multisports,cui2023sportsmot,wu2024sportshhi}---soccer has become an important domain for fine-grained video analysis and commentary generation, supported by recent soccer datasets and models~\cite{gao2023trajectory,cioppa2020multimodal,cioppa2024soccernet,giancola2018soccernet_action_spotting,qi2023goal,rao2025towards,rao2024matchtime,mkhallati2023soccernet}.
In this work, we focus on real-time streaming soccer commentary, where a model must provide low-latency narration over an untrimmed full match.
Unlike generic video captioning~\cite{chen2011collecting,xu2016msr,krishna2017dense,zhou2018towards,wang2019vatex}, soccer commentary is a domain-specific narration task that requires professional event understanding and entity grounding~\cite{mkhallati2023soccernet,qi2023goal,rao2024matchtime}.
A useful streaming commentator should respond with low latency, mention key events at the right time, use correct player and team names, and avoid factually incorrect descriptions.

Existing soccer commentary models can be roughly grouped into offline and streaming settings, as shown in Fig.~\ref{fig:issues}.
Offline soccer commentary models~\cite{qi2023goal,rao2024matchtime,rao2025towards,gautam2025soccerchat,li2025soccercomment,you2025timesoccer} are usually designed for pre-segmented clips or localized events.
When applied to uncut full-match videos with sliding windows, they are not naturally synchronized with the live timeline and may produce delayed, repeated, or poorly timed commentary.
Recent streaming models~\cite{chen2025livecc,xu2025streamingvlm} make real-time full-match commentary feasible by using timestamp-aligned speech supervision, interleaved video-text generation, and streaming inference mechanisms such as sliding windows or KV-cache reuse.
However, these models mainly learn to speak continuously and fluently, and may still generate generic descriptions without clearly covering important soccer events.
For example, they can describe nearby player movements or referee actions, but fail to explicitly mention whether the moment is a goal, card, substitution, or offside.
Prior soccer commentary benchmarks suggest that professional soccer commentary should be factual, temporally aligned, and aware of fine-grained entities, actions, and event semantics~\cite{qi2023goal,ge2024scbench,mkhallati2023soccernet,cioppa2024soccernet,gautam2024soccernet_echoes,rao2024matchtime}.
Therefore, we focus on event-grounded streaming soccer commentary: improving the model's ability to mention key events accurately and in time while preserving the low-latency streaming protocol.

Following recent streaming commentary models~\cite{chen2025livecc,xu2025streamingvlm}, the model processes incoming frames as an interleaved video-text stream and generates only a few words per second, mimicking the pace of human commentary.
During SFT, each second is supervised by the aligned text increment, with a pause token such as ``...'' for silent or incomplete moments.
With bounded sliding windows and KV-cache reuse, this interface enables real-time narration over untrimmed full-match videos.

However, human-paced text increments per second are often too short to carry complete key-event descriptions.
As shown in Fig.~\ref{fig:supervision_granularity_mismatch}, one-second ASR fragments rarely indicate the target event, while reliable event coverage usually requires commentary over multiple seconds.
This motivates defining event-level feedback over macro-window rollouts while keeping inference at the original low-latency streaming granularity.
To make this feasible, the model also needs reliable time-anchored supervision from noisy broadcast ASR, and evaluation should measure factual and temporal event grounding rather than only surface fluency.
We therefore develop \textbf{SoccerNarrate}, a unified framework that couples time-anchored data construction, macro-window preference alignment, and event-centric evaluation. Our main contributions are summarized as follows:
\begin{itemize}
    \item \textbf{Time-anchored data for streaming soccer commentary.}
    We construct \textbf{SoccerNarrate-Data}, a large-scale dataset for full-match streaming soccer commentary.
    It is built with word-level ASR alignment, context-aware quality filtering, and phoneme-based entity calibration, providing dense supervision for streaming SFT and preference construction.

    \item \textbf{Macro-window preference alignment for event-grounded streaming.}
    We perform macro-window preference alignment (MWPA) to align streaming commentary models with complete event-level commentary semantics.
    MWPA defines preferences over causal multi-step rollouts from the same streaming prefix rather than isolated one-second outputs, while keeping second-level inference unchanged.
    We instantiate this alignment with Direct Preference Optimization (DPO) and use event-mismatched human commentaries as counterfactual negatives, encouraging event semantics over commentary style alone.

    \item \textbf{Event-centric evaluation with temporal matching.}
    We introduce \textbf{SoccerNarrate-Eval}, a full-match benchmark for evaluating whether generated commentary correctly mentions key soccer events under temporal constraints.
    We formulate evaluation as event entailment detection with temporally constrained event matching, and show that SoccerNarrate improves event coverage and event-aligned precision over offline and streaming baselines.
\end{itemize}

\setcounter{figure}{1}
\begin{figure}
    \centering
    \includegraphics[width=\linewidth]{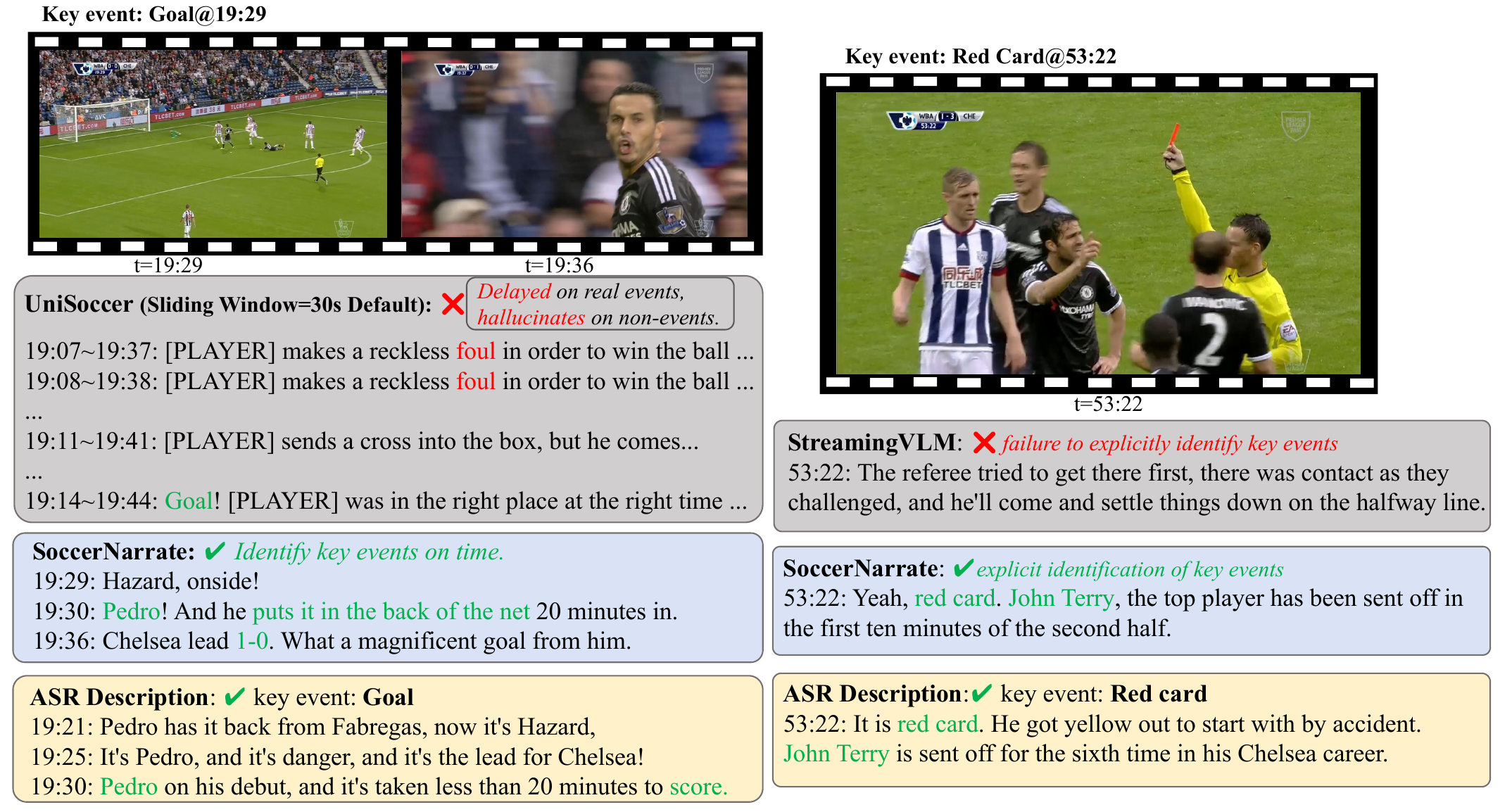}
    \caption{\textbf{Issues in existing soccer commentary models.}
 \textbf{Left:} Offline models react late to real events and may also produce hallucinated commentary on non-events. \textbf{Right:} Streaming/online models describe the surrounding context but do not clearly recognize the key event. In contrast, SoccerNarrate produces timely and event-grounded commentary.}
    \label{fig:issues}
\end{figure}

\begin{figure}[t]
    \centering
    \includegraphics[width=0.68\linewidth]{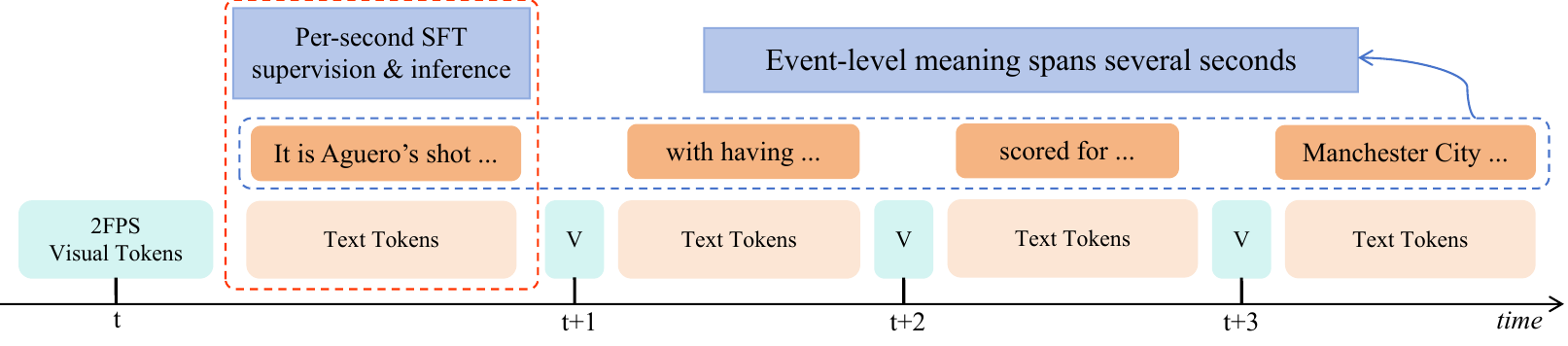}
    \hfill
    \includegraphics[width=0.3\linewidth]{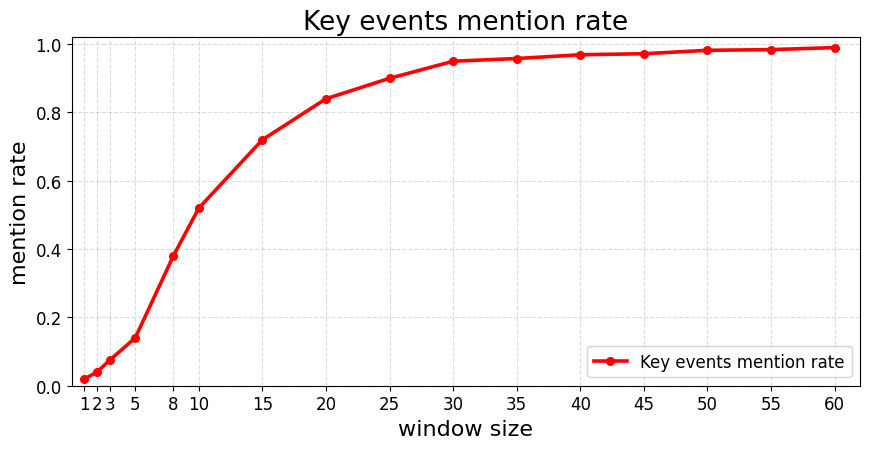}
    \caption{
\textbf{Motivation for macro-window event feedback.} 
Streaming models generate and are supervised at second-level steps, while explicit mentions of key soccer events in human commentary often appear across multiple seconds. The right plot measures whether ASR text centered on ground-truth events entails the corresponding event. 
One-second fragments rarely contain explicit event mentions, while longer windows provide a more reliable unit for event-level preference construction and evaluation.
}
    \label{fig:supervision_granularity_mismatch}
\end{figure}

\label{sec:motivation}

\section{Related work}

\paragraph{Soccer commentary datasets and benchmarks.}
Existing soccer commentary resources differ substantially in scale, temporal granularity, and supervision reliability, as summarized in Table~\ref{tab:datasets}. 
Clip- or sentence-level datasets~\cite{mkhallati2023soccernet,rao2025towards,qi2023goal,ge2024scbench,rao2024matchtime} support soccer captioning generation, but their annotations are usually provided as clip-centered anchors, localized event comments, or sentence-level timestamps, making them less directly suited to second-level streaming supervision. 
ASR-based resources~\cite{gautam2024soccernet_echoes,chen2025livecc,xu2025streamingvlm} provide denser time-aligned speech supervision; however, ASR-derived transcripts can still contain transcription noise and entity errors, and existing sports-streaming datasets are not specifically curated for soccer entity grounding and event-level evaluation. 
On the benchmark side, prior evaluations mainly rely on offline captioning metrics, LLM-based scoring, or pairwise LLM-as-a-judge win rates, which capture fluency and general preference but do not directly measure whether key soccer events are mentioned correctly under temporal constraints. 
In contrast, SoccerNarrate targets this gap by providing full-match word-level text streams with post-processing and entity calibration, and by supporting event-centric evaluation through temporally constrained event matching.

\paragraph{VLMs for soccer commentary and streaming understanding.}
Recent VLM-based soccer commentary systems mainly follow two directions.
Offline soccer-specific models~\cite{rao2024matchtime,rao2025towards,gautam2025soccerchat,li2025soccercomment} generate commentaries from pre-segmented clips or localized events, improving soccer-domain captioning but not directly supporting causal full-match streaming.
When deployed with sliding windows, they are not naturally synchronized with the live stream and may produce delayed, repeated, or poorly timed descriptions.
TimeSoccer~\cite{you2025timesoccer} instead processes a full 45-minute half end-to-end and jointly predicts timestamps and commentary, but it observes the entire half before generating and thus cannot narrate causally.
Perception-oriented soccer foundation models such as SoccerMaster~\cite{yang2026soccermaster} instead learn general soccer representations through multi-task training on pre-trimmed clips, which is complementary to generative commentary that must decide whether, when, and how to describe events during causal full-match inference.
In contrast, online VLMs~\cite{chen2024videollm,huang2025online,yang2025livestar} focus on streaming memory, context compression, and response timing for continuous video understanding, but are not specifically designed for soccer event grounding.
The closest works, LiveCC~\cite{chen2025livecc} and StreamingVLM~\cite{xu2025streamingvlm}, use timestamp-aligned ASR supervision to enable low-latency sports commentary over long streams.
However, their training is still primarily based on second-level generation targets, while event-level correctness in soccer commentary is more reliably assessed over multi-second contexts.
Our work builds on this streaming capability and introduces macro-window preference alignment to improve event-grounded soccer commentary under low-latency inference.

\paragraph{Granularity-aware preference optimization.}
Direct Preference Optimization (DPO)~\cite{rafailov2023direct} aligns language models using preference pairs without explicit reward modeling. 
Recent variants suggest that preference optimization can benefit from choosing feedback units that match the semantic granularity of the target behavior. 
For example, SDPO~\cite{kong2025sdpo} selects key segments across multi-turn agent interactions, where each turn is already a complete response, while ASPO~\cite{wang2025aspo} and Mask-DPO~\cite{gu2025mask} assign sentence-level or factuality-aware credit within a complete response.
Video-DPO~\cite{liu2025videodpo} further extends preference learning to multimodal and video settings, but typically assumes offline inputs where the model can access complete visual evidence before generating a response.
In streaming soccer commentary, however, the response itself is fragmented into short text increments interleaved with newly arriving visual tokens and silent steps, so a single step rarely forms a complete event description.
We therefore group these causal emissions into a macro-window trajectory from the same streaming prefix, and contrast it with event-mismatched human commentary to provide a preference signal that is more focused on event grounding than on generic fluency, verbosity, or surface commentary style.
\section{Method}

In this section, we introduce the SoccerNarrate framework.
We first describe SoccerNarrate-Data, which provides time-anchored word-level supervision for streaming commentary.
We then present a DPO-based macro-window alignment strategy that applies event-level preference feedback to causal multi-step rollouts while keeping second-level streaming inference unchanged.
Finally, we define the event-centric evaluation protocol used to measure whether generated commentary mentions key soccer events correctly under temporal constraints.
\subsection{SoccerNarrate-Data}

\paragraph{Curation.}
We construct SoccerNarrate-Data from publicly available SoccerNet~\cite{cioppa2024soccernet} broadcast matches.
Specifically, we use 436 full-match videos with commentary audio, where each match lasts about 1.5 hours.
Following the official SoccerNet split, 391 matches are used for training and 45 matches are held out for evaluation.
Given a full-match video, our goal is to convert raw broadcast audio into a clean, time-anchored text stream for streaming narration.

\begin{figure}[t]
  \centering
  \includegraphics[width=0.95\textwidth]{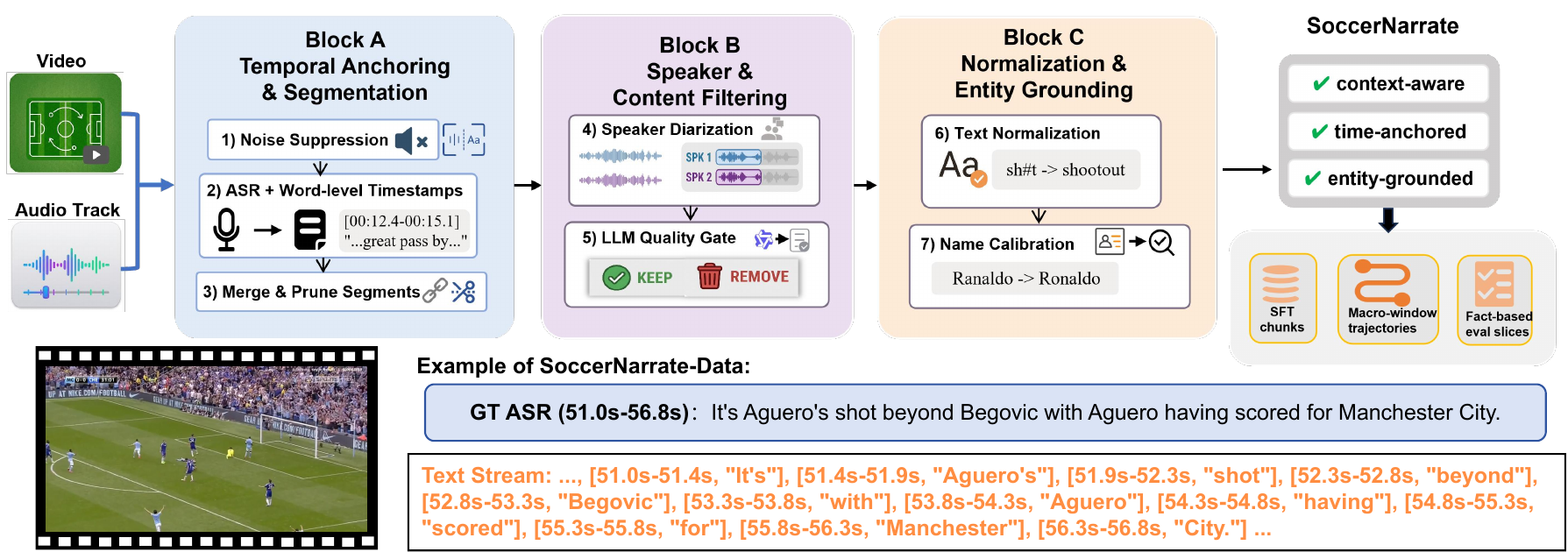}
  \caption{\textbf{SoccerNarrate-Data curation pipeline.}
Starting from SoccerNet videos, we extract word-level ASR transcripts from match audio, filter noisy or irrelevant commentary, and normalize entity mentions using match metadata.
The output is a clean, time-anchored, and entity-grounded text stream for streaming SFT, macro-window preference construction, and event-centric evaluation.
The bottom example shows how a commentary sentence is converted into timestamped word tokens.}
  \label{fig:pipeline_and_example}
\end{figure}

As shown in Fig.~\ref{fig:pipeline_and_example}, the pipeline contains three stages.
First, we suppress background noise and run WhisperX-large-v3~\cite{bain2023whisperx} to obtain ASR transcripts with word-level timestamps, followed by merging and pruning fragmented segments.
Second, we perform speaker diarization with pyannote.audio to identify commentator speech, and apply a context-aware LLM quality gate based on Qwen3-235B with multi-round voting to remove noisy, irrelevant, or non-commentary text.
Third, we normalize the text and calibrate player/team names by aligning LLM-extracted entity mentions to match rosters using phoneme-level similarity.
The final output is a word-level ASR text stream, as illustrated in Fig.~\ref{fig:pipeline_and_example}, which supports streaming SFT, macro-window preference construction, and temporally grounded event-centric evaluation.

\paragraph{Statistics.}
Table~\ref{tab:datasets} compares SoccerNarrate-Data with existing soccer captioning and ASR datasets.
Unlike clip-centered caption datasets, SoccerNarrate-Data provides dense word-level timestamps for full-match streaming supervision.
Compared with existing ASR-based soccer data, it preserves broad text coverage while substantially reducing transcription noise and entity errors.
We report the number of games, temporal granularity, post-processing, ASR-covered duration, total words, commentary segments, and name error rate. We validate the curation quality through human checks on sampled matches, including timestamp alignment, spelling, entity calibration, and event-judgment consistency; details are provided in Appendix~\ref{app:data_validation}.

\begin{table}[t]
\caption{\textbf{Comparison with existing soccer captioning and ASR datasets.}
SoccerNarrate-Data provides full-match word-level text streams with post-processing and entity calibration, enabling dense and reliable supervision for streaming soccer commentary.}
\label{tab:datasets}
\centering
\setlength{\tabcolsep}{3.0pt}
\definecolor{groupgray}{gray}{0.92}
\resizebox{0.95\columnwidth}{!}{
\begin{tabular}{lccccccc}
\toprule
\textbf{Dataset} 
& \textbf{Game} 
& \textbf{Time Gran.} 
& \textbf{Post-proc.} 
& \textbf{Text Dur. (h)} 
& \textbf{Words (k)} 
& \textbf{Comm. (k)} 
& \textbf{Name Err. (\%)} \\
\midrule

\rowcolor{groupgray}
\multicolumn{8}{l}{\textit{Clip-centered soccer captioning datasets}} \\
SoccerNet-Caption~\cite{mkhallati2023soccernet}
& 471 & Clip-center & \xmarklight & 267 & 855 & 36.8 & - \\
SoccerReplay~\cite{rao2025towards}
& 1988 & Clip-center & \cmark & 1025 & 3145 & 141.9 & - \\

\midrule
\rowcolor{groupgray}
\multicolumn{8}{l}{\textit{Sentence-level commentary and ASR datasets}} \\
GOAL~\cite{qi2023goal}
& 20 & Sentence & \xmarklight & 21 & 193 & 8.9 & - \\
SCBench (Soccer)~\cite{ge2024scbench}
& - & Sentence & \cmark & 2 & 14 & 0.7 & - \\
SoccerNet-Echoes~\cite{gautam2024soccernet_echoes}
& 550 & Sentence & \xmarklight & 569 & 4760 & 53.8 & 28.4 \\

\midrule
\rowcolor{groupgray}
\multicolumn{8}{l}{\textit{Word-level streaming ASR datasets}} \\
LiveSports (Soccer)~\cite{chen2025livecc}
& - & Word & \cmark & 21 & 217 & 0.9 & - \\
Inf-Stream (Soccer)~\cite{xu2025streamingvlm}
& 443 & Word & \cmark & 176 & 1820 & 45.4 & 0.4 \\

\midrule
\rowcolor{groupgray}
\multicolumn{8}{l}{\textit{Ours}} \\
\textbf{SoccerNarrate-Data}
& \textbf{436} & \textbf{Word} & \cmark 
& \textbf{502} & \textbf{4657} & \textbf{58.6} & \textbf{0.1} \\

\bottomrule
\end{tabular}
}
\end{table}

\subsection{Macro-window Preference Alignment under Streaming Inference}
\label{sec:mws_dpo}

\paragraph{Streaming commentary task.}
Following LiveCC~\cite{chen2025livecc} and StreamingVLM~\cite{xu2025streamingvlm}, we formulate online soccer commentary as causal generation over a continuously arriving video stream.
The match is sampled at 2 FPS and grouped into 1-second steps.
At step $t$, the model receives the newly arrived visual tokens $V_t$ and the previous text tokens $W_{<t}$, and predicts the commentary increment $W_t$:
\begin{equation}
    W_t \sim \pi_\theta(\cdot \mid V_{\leq t}, W_{<t}),
\end{equation}
where $V_{\leq t}$ denotes all visual evidence available up to the current step, $W_{<t}$ denotes previously generated or teacher-forced commentary tokens, and $\pi_\theta$ is the streaming policy model.
The target $W_t$ is obtained from the timestamped ASR text stream during training, and generated autoregressively during inference.

For long-match inference, we follow the StreamingVLM cache setting to bound memory and latency.
Specifically, the model keeps 512 attention-sink tokens, the most recent 512 text tokens, and the latest 16 seconds of visual tokens, corresponding to 32 frames at 2 FPS.
Older visual tokens are evicted when the cache exceeds this budget, while sink and recent text tokens are retained for stable streaming generation.

\paragraph{DPO over causal macro-window rollouts.}
The model generates commentary at 1-second streaming steps, but each step usually contains only a short text increment or silence.
Such local outputs are often too fragmented to serve as a reliable unit for event-level preference comparison.
We therefore keep the second-level streaming inference process unchanged, and instantiate DPO over an $L$-step macro-window rollout.

Starting from the same streaming prefix state at step $k$, we compare two $L$-step commentary continuations:
\begin{equation}
    Y_k = (W_k, W_{k+1}, \ldots, W_{k+L-1}),
\end{equation}
where $W_t$ denotes the commentary increment generated at streaming step $t$.
For a preference pair, $Y_k^{+}$ and $Y_k^{-}$ denote the preferred and rejected trajectories, respectively.
We denote the shared training condition as
\begin{equation}
    C_k = (s_{k-1}, V_{k:k+L-1}),
\end{equation}
where $s_{k-1}$ is the shared streaming prefix state before step $k$, and $V_{k:k+L-1}$ is the video stream that arrives during the macro-window rollout.
This notation does not imply non-causal access to future frames: when scoring tokens at step $t$, the model only conditions on the prefix state and visual-text context available up to that step.

We use the standard DPO objective, but apply it to macro-window streaming trajectories:
\begin{equation}
\mathcal{L}_{\mathrm{MWPA}}
=
-\mathbb{E}_{(C_k,Y_k^+,Y_k^-)}
\log \sigma
\left(
\beta
\left[
\log \frac{\pi_\theta(Y_k^+ \mid C_k)}
         {\pi_\theta(Y_k^- \mid C_k)}
-
\log \frac{\pi_{\mathrm{ref}}(Y_k^+ \mid C_k)}
         {\pi_{\mathrm{ref}}(Y_k^- \mid C_k)}
\right]
\right),
\end{equation}
where $\pi_\theta$ is the trainable streaming policy, $\pi_{\mathrm{ref}}$ is a frozen reference model, and $\beta$ controls the preference strength.
The trajectory log-probability is computed by summing token log-probabilities over all text tokens in the macro window, under the same causal streaming mask used for generation.
In this way, macro-window preference alignment provides event-level feedback while leaving the second-level streaming inference process unchanged.

\begin{figure}
    \centering
    \includegraphics[width=0.98\linewidth]{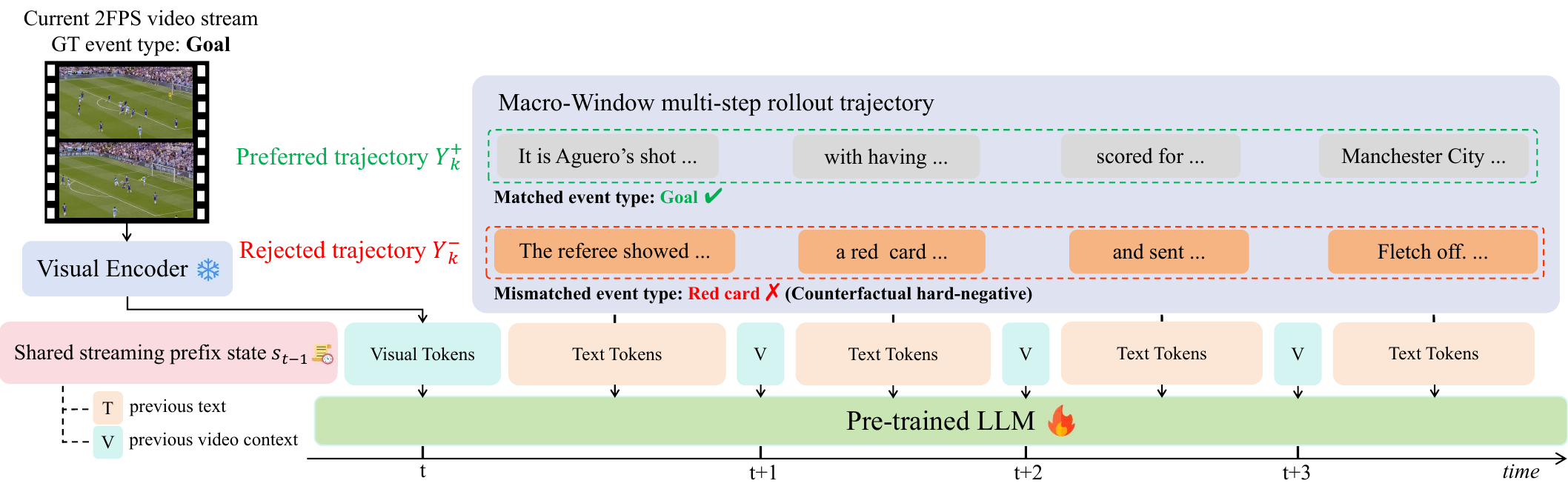}
    \caption{
\textbf{Overview of Macro-window Preference Alignment (MWPA) and counterfactual hard-negative construction.}
Given the same streaming prefix and video window, MWPA compares two human commentary trajectories from the same match.
The preferred trajectory $Y_k^{+}$ matches the target event, while the rejected trajectory $Y_k^{-}$ is sampled from a type-mismatched event as a counterfactual hard negative.
This provides event-level preference supervision while preserving micro-step streaming inference.
}
    \label{fig:mws_dpo}
\end{figure}

\paragraph{Counterfactual hard-negative construction.}
We build preference pairs from SoccerNet(train-split) event annotations and our timestamped ASR commentary stream.
SoccerNet-label~\cite{cioppa2024soccernet} provides ground-truth timestamps and event types for key soccer events, covering 17 event classes.
For each annotated target event $(t,c)$, we take the ASR commentary within a fixed macro window around $t$ as the preferred trajectory $Y_k^+$.
The rejected trajectory $Y_k^-$ is randomly sampled from another event window in the same match but with a different event type $c' \neq c$.
Thus, both $Y_k^+$ and $Y_k^-$ are human commentaries from the same game, which helps control match context, teams, players, and commentator style, while making the rejected trajectory visually unsupported and semantically mismatched under the target window.

To reduce class imbalance, we balance the number of preference pairs across event categories.
To avoid preference decisions being dominated by very sparse commentary, we only keep windows with a word density above 1.5 words per second; the retained samples contain about 2 words per second on average.
This process yields around 15K macro-window preference pairs for MWPA training.

\subsection{Benchmark protocol \& metrics}
\textbf{Event-centric evaluation.} Generic captioning metrics and pairwise win rates do not directly measure whether a streaming commentator mentions key soccer events correctly and on time.
A fluent commentary can still miss a goal, mention a card too late, or hallucinate a substitution.
We therefore build SoccerNarrate-Eval as an event-centric benchmark based on temporally constrained event entailment. On the 45 SoccerNarrate-Eval matches, we use SoccerNet-label event annotations as ground truth.
We slide a 30-second window over the generated commentary with a 15-second stride, and use Qwen3-235B to detect whether each window entails any target event category. A detected event in window $[\hat{s}, \hat{e}]$ with class $\hat{c}$ is counted as correct if it matches a ground-truth event $(t, c)$ with the same class, $\hat{c}=c$, and the temporal distance from $t$ to the window interval is no larger than $\delta=15$ seconds, i.e.,
$d(t,[\hat{s},\hat{e}]) \le 15$.
The distance is zero when $t \in [\hat{s},\hat{e}]$ and otherwise equals the distance to the nearest window boundary.
We enforce one-to-one matching to avoid counting repeated mentions as multiple correct hits.
We report event-level Recall, Precision, and F1 over 14 evaluated event classes.
Recall measures event coverage, while Precision penalizes hallucinated, duplicated, or poorly timed event mentions.
We exclude \textsc{Kick-off}, \textsc{Throw-in}, and \textsc{Clearance}, since they are frequent routine actions and are rarely verbalized explicitly in live commentary. Details are provided in Appendix~\ref{app:details_about_SoccerNarrate}.

\textbf{Commentary latency evaluation.} We evaluate commentary latency along two complementary dimensions: \emph{temporal mention delay} and \emph{streaming inference latency}.
First, mention delay relative to the ground-truth event timestamp $t$ is directly bounded by the temporal matching tolerance $\delta$ in SoccerNarrate-Eval: mentions arriving outside $d(t,[\hat{s},\hat{e}]) \le \delta$ fail to match and penalize both Recall and Precision, while sweeping $\delta$ from $0$\,s to $60$\,s (Appendix~\ref{app:eval_protocol_robustness}) further profiles sensitivity to delayed or retrospective narration.
Second, to verify real-time responsiveness during continuous full-match decoding, we follow StreamingVLM~\cite{xu2025streamingvlm} and measure per-token generation latency (s/token) over a 45-minute half against the real-time threshold of 10 tokens/s (0.1\,s/token), assessing both average speed and long-horizon stability (Appendix~\ref{app:latency_analysis}, Fig.~\ref{fig:latency_appendix}).
\section{Experiments}

\begin{table*}[t]
\caption{
\textbf{Results on SoccerNarrate-Eval.}
We report class-averaged and overall Recall, Precision, and F1 Score on the 45 test matches under the same event-centric protocol.
Offline models are applied to full matches with non-overlapping 30-second windows, where \#Frames is the number of frames per window; TimeSoccer instead takes each 45-minute half as input with 288 uniformly sampled frames.
Streaming models run causally over the full match at the listed frame rate, and (FT) denotes fine-tuning on SoccerNarrate-SFT.
Detailed per-class breakdown is deferred to the Appendix~\ref{app:per_class}.
}
\label{tab:main_compact_groups}
\centering
\setlength{\tabcolsep}{5.2pt}
\definecolor{sectiongray}{gray}{0.92}
\definecolor{algogray}{gray}{0.98}

\resizebox{0.95\linewidth}{!}{
\small
\begin{tabular}{lcccccccc}
\toprule
\multirow{2}{*}{\textbf{Model}} & \multirow{2}{*}{\textbf{Size}} & \multirow{2}{*}{\textbf{\#Frames}}
& \multicolumn{2}{c}{\textbf{Recall (\%)}} 
& \multicolumn{2}{c}{\textbf{Precision (\%)}} 
& \multicolumn{2}{c}{\textbf{F1 (\%)}} \\
\cmidrule(lr){4-5} \cmidrule(lr){6-7} \cmidrule(lr){8-9}
& & 
& \textit{Class Avg.} & \textit{Overall Avg.}
& \textit{Class Avg.} & \textit{Overall Avg.}
& \textit{Class Avg.} & \textit{Overall Avg.} \\
\midrule

\rowcolor{sectiongray}
\multicolumn{9}{l}{\textit{General offline MLLMs}} \\

Video-LLaVA~\cite{lin2024video_llava} & 7B & 60 & 14.5 & 11.2 & 14.0 & 8.1 & 4.0 & 9.4 \\

Qwen2.5-VL~\cite{bai2025qwen2} & 7B & 60 & 32.6 & 18.8 & 28.1 & 31.1 & 22.6 & 23.4 \\

Qwen3-VL~\cite{yang2025qwen3} & 8B & 60& 19.5 & 19.4 & 33.2 & 38.8 & 19.8 & 25.8 \\

Qwen3.5~\cite{team2026qwen3_5} & 9B & 60& 42.6 & 52.9 & 19.5 & 28.8 & 23.8 & 37.3 \\

\midrule
\rowcolor{sectiongray}
\multicolumn{9}{l}{\textit{Offline soccer commentary models}} \\

MatchTime~\cite{rao2024matchtime}& 8B & 30 & 11.8 & 21.7 & 10.7 & 23.7 & 9.6 & 22.7 \\

UniSoccer~\cite{rao2025towards}
& 8B & 30 & 50.4 & 50.0 & 28.5 & 23.5 & 29.4 & 32.0 \\

SoccerChat~\cite{gautam2025soccerchat}
& 7B & 30 & 51.1 & 63.5 & 26.5 & 35.6 & 32.9 & 45.6 \\

TimeSoccer~\cite{you2025timesoccer}
& 7B & 288 & 41.5 & 42.9 & 24.1 & 23.6 & 24.5 & 30.4 \\

\midrule
\rowcolor{sectiongray}
\multicolumn{9}{l}{\textit{General streaming models}} \\

VideoChat-Online~\cite{huang2025online}
& 4B & 2 fps & 3.2 & 1.6 & 13.9 & 9.4 & 3.0 & 2.8 \\

LiveCC~\cite{chen2025livecc}
& 7B & 2 fps & 2.6 & 2.4 & 11.7 & 7.9 & 2.7 & 3.7 \\

LiveStar~\cite{yang2025livestar}
& 8B & 1 fps & 17.5 & 12.4 & 22.0 & 24.2 & 11.9 & 16.4 \\

\midrule
\rowcolor{sectiongray}
\multicolumn{9}{l}{\textit{Streaming soccer commentary models}} \\

LiveCC (FT)
& 7B & 2 fps & 8.2 & 6.4 & 27.1 & 34.0 & 9.7 & 10.7 \\

StreamingVLM~\cite{xu2025streamingvlm}
& 7B & 2 fps & 33.2 & 29.4 & 30.3 & 40.9 & 28.5 & 34.2 \\

StreamingVLM (FT)
& 7B & 2 fps & 34.6 & 30.3 & 31.5 & 42.3 & 30.0 & 35.3 \\
\midrule
\rowcolor{algogray}
\textbf{SoccerNarrate}
& 7B & 2 fps & \textbf{71.8} & \textbf{61.3} & \textbf{38.5} & \textbf{48.8} & \textbf{46.9} & \textbf{54.4} \\

\bottomrule
\end{tabular}
}
\end{table*}

\subsection{Experimental Setup}

For a fair comparison with StreamingVLM, we initialize from the same backbone, Qwen2.5-VL-Instruct-7B~\cite{bai2025qwen2}.
Training proceeds in three stages.
\textbf{Stage~1 (streaming-format warm-up)} trains on StreamingVLM-SFT-525K~\cite{xu2025streamingvlm} and Live-WhisperX-526K~\cite{chen2025livecc} (1.051M samples total) to learn streaming generation.
\textbf{Stage~2 (soccer-domain SFT)} further fine-tunes on our 58K SoccerNarrate-SFT dataset to better match professional live commentary in both style and factual narration.
\textbf{Stage~3 (preference optimization)} applies MWPA on SoccerNarrate-Pref, our 15K preference pairs, to align model outputs with human preferences. The total compute is about 136 H100-days.

\subsection{Main Results}

\paragraph{Overall comparison.}
Table~\ref{tab:main_compact_groups} compares offline, streaming, and soccer-specialized models on SoccerNarrate-Eval; detailed per-class results are provided in Appendix~\ref{app:per_class} and per-league results in Appendix~\ref{app:per_league}.
First, additional soccer-domain second-level SFT brings limited event-level gains.
For example, LiveCC improves from 2.7 to 9.7 in class-average F1 after fine-tuning on 58K SoccerNarrate-SFT samples, while StreamingVLM only improves from 28.5 to 30.0.
This suggests that second-level supervision helps domain adaptation, but is not always sufficient for optimizing key-event correctness.

Second, macro-window preference alignment consistently improves Recall, Precision, and F1.
Compared with StreamingVLM (FT), SoccerNarrate improves class/overall F1 from 30.0/35.3 to 46.9/54.4.
Since the second-level streaming inference protocol is unchanged, these gains indicate that applying preferences over causal multi-step rollouts provides useful event-level feedback.

Third, offline soccer commentary models can achieve high event recall, but when deployed on full matches using non-overlapping 30-second sliding windows, their precision is much lower.
SoccerChat reaches a comparable overall Recall to SoccerNarrate (63.5 vs.\ 61.3), but its class/overall Precision is lower by 12.0/13.2 points.
This suggests that clip-centered offline models can mention salient events, but are less reliable in deciding when to trigger event descriptions during continuous inference.
Overall, SoccerNarrate improves event coverage while maintaining better event-aligned precision under low-latency streaming inference.

\paragraph{Win-rate comparison.}
We further report win-rate judged by GPT-4o on Inf-Streams-Eval (110 ASR segments from 4 soccer matches) and SoccerNarrate-Eval (7851 ASR segments from the 45 full matches).
As shown in Table~\ref{tab:win_rate}, SoccerNarrate clearly outperforms LiveCC and StreamingVLM on SN-Eval and remains competitive on IS-Eval, suggesting that the event-level gains do not come at the cost of holistic commentary preference.

\paragraph{SCBench commentary quality.}
We also evaluate on the soccer subset of SCBench~\cite{ge2024scbench} using SCORES, as shown in Table~\ref{tab:scores_scbench_abbrev}.
Each clip is annotated with one to three relevant dimensions, on which GPT-4o scores the prediction against the reference commentary from 0 to 10; scores are averaged per dimension and over all soccer clips.
SoccerNarrate achieves the best average score (2.38), with consistent gains across all six commentary dimensions.
This provides complementary evidence on broader commentary quality beyond the event-centric metrics.

\begin{table*}[t]
\centering

\begin{minipage}[t]{0.42\textwidth}
\centering
\caption{
Win Rate A vs. B comparison on different benchmarks, judged by GPT-4o.
IS-Eval = Inf-Streams-Eval, SN-Eval = SoccerNarrate-Eval.
}
\label{tab:win_rate}
\vspace{-0.6em}
\footnotesize
\setlength{\tabcolsep}{0.3pt}
\renewcommand{\arraystretch}{0.9}
\begin{tabular}{lcccc} 
\toprule
\multicolumn{1}{c}{\textbf{Win Rate A vs. B}} 
& \multicolumn{2}{c}{\textbf{IS-Eval (Soccer)}} 
& \multicolumn{2}{c}{\textbf{SN-Eval}} \\
\cmidrule(lr){2-3} \cmidrule(lr){4-5} 
\diagbox[width=7em]{\textbf{Model A}}{\textbf{Model B}} 
& \textbf{LiveCC} & \textbf{SVLM} 
& \textbf{LiveCC} & \textbf{SVLM} \\
\midrule
Qwen2.5-VL~\cite{bai2025qwen2} & 86.2 & 30.6 & 81.2 & 25.1 \\
LiveCC~\cite{chen2025livecc} & -- & 14.7 & -- & 10.0 \\
SVLM~\cite{xu2025streamingvlm} & 85.3 & -- & 90.0 & -- \\
\textbf{SoccerNarrate} & 67.3 & 50.2 & 96.2 & 78.5 \\
\bottomrule
\end{tabular}
\end{minipage}
\hfill
\begin{minipage}[t]{0.56\textwidth}
\centering
\caption{
\textbf{SCORES (0--10) on commentary quality dimensions} evaluated on the soccer subset of SCBench. \textbf{K}=Key Event Caption, \textbf{T}=Technical Detail Analysis, \textbf{B}=Background Information Interpretation, \textbf{C}=Tactical Analysis, \textbf{M}=Match Situation Interpretation, \textbf{E}=Emotional Expression.
}
\label{tab:scores_scbench_abbrev}
\vspace{-0.6em}
\footnotesize
\setlength{\tabcolsep}{3.2pt}
\renewcommand{\arraystretch}{0.95}
\begin{tabular}{lccccccc}
\toprule
\textbf{Model} 
& \textbf{K} & \textbf{T} & \textbf{B} & \textbf{C} & \textbf{M} & \textbf{E} & \textbf{Avg.} \\
\midrule
LiveCC~\cite{chen2025livecc} & 2.38 & 1.69 & 1.38 & 1.20 & 1.65 & 1.49 & 1.65 \\
StreamingVLM~\cite{xu2025streamingvlm} & 2.60 & 1.59 & 1.32 & 0.73 & 1.09 & 1.46 & 1.46 \\
\textbf{SoccerNarrate} & 2.63 & 2.24 & 2.38 & 2.37 & 2.31 & 2.60 & 2.38 \\
\bottomrule
\end{tabular}
\end{minipage}

\vspace{-0.6em}
\end{table*}

\subsection{Ablations}
\paragraph{Effect of macro-window size.}
Table~\ref{tab:window_size_ablation} shows that MWPA benefits from an appropriate rollout length.
The 1s setting remains close to SFT, indicating that isolated streaming increments are too fragmented for event-level preference comparison.
Performance improves from 1s to 15s, confirming that multi-step rollouts provide a more reliable alignment unit.
However, 20--40s windows do not yield consistent gains, likely because unrelated commentary dilutes the target-event signal.
Although 10s obtains slightly higher overall F1 due to frequent classes, 15s achieves the best class-averaged F1 and precision, so we use it as the default.

\paragraph{Effect of preference-data scale.}
Table~\ref{tab:ablation_pref_data_scaling} shows that increasing MWPA preference data steadily improves averaged Recall and F1.
Class-averaged F1 rises from 30.0 to 46.9, and overall F1 from 35.3 to 54.4.
The larger gain in Recall suggests that broader event-mismatched supervision mainly improves event coverage while preserving event-aligned precision.
\begin{table}[t]
\centering
\begin{minipage}[t]{0.45\linewidth}
\centering
\scriptsize
\caption{
\textbf{Ablation study on macro-window size.}
The SFT baseline uses no preference window.
}
\label{tab:window_size_ablation}
\setlength{\tabcolsep}{4pt}
\renewcommand{\arraystretch}{0.98}
\resizebox{\linewidth}{!}{
\begin{tabular}{l c ccc ccc}
\toprule
\multirow{2}{*}{\textbf{Method}} 
& \multirow{2}{*}{\textbf{Win.}} 
& \multicolumn{3}{c}{\textbf{Class Avg.}} 
& \multicolumn{3}{c}{\textbf{Overall Avg.}} \\
\cmidrule(lr){3-5} \cmidrule(lr){6-8}
& 
& \textbf{R} & \textbf{P} & \textbf{F1}
& \textbf{R} & \textbf{P} & \textbf{F1} \\
\midrule
SFT & -- 
& 34.6 & 31.5 & 30.0 
& 30.3 & 42.3 & 35.3 \\
\midrule
\multirow{9}{*}{+MWPA}
& 1 s  
& 34.0 & 30.9 & 30.2 
& 28.4 & 41.7 & 33.8 \\
& 5 s  
& 51.2 & 31.1 & 32.3 
& 47.7 & 42.6 & 45.0 \\
& 10 s 
& 66.5 & 34.0 & 42.5 
& \textbf{66.8} & 48.2 & \textbf{56.0} \\
& \textbf{15 s} 
& 71.8 & \textbf{38.5} & \textbf{46.9} 
& 61.3 & \textbf{48.8} & 54.4 \\
& 20 s 
& \textbf{72.2} & 37.3 & 45.1 
& 59.1 & 48.7 & 53.4 \\
& 25 s 
& 68.9 & 35.6 & 43.3 
& 60.2 & 48.1 & 53.5 \\
& 30 s 
& 70.8 & 32.0 & 39.7 
& 60.0 & 44.8 & 51.3 \\
& 35 s 
& 69.1 & 36.3 & 43.1 
& 56.6 & 48.4 & 52.2 \\
& 40 s 
& 68.4 & 35.6 & 42.8 
& 56.6 & 47.6 & 51.7 \\
\bottomrule
\end{tabular}
}
\end{minipage}
\hfill
\begin{minipage}[t]{0.52\linewidth}
\centering
\small
\caption{
\textbf{Ablation study on preference-data scaling.}
We report average metrics using different amounts of MWPA preference data under the setting of 15-second macro window. Per-event scaling results are provided in Appendix~\ref{app:per_event_mwpa_data_scaling}.
}
\label{tab:preference_data_scaling_avg}
\setlength{\tabcolsep}{2.8pt}
\renewcommand{\arraystretch}{1.05}
\resizebox{\linewidth}{!}{
\begin{tabular}{l l ccc ccc}
\toprule
\multirow{2}{*}{\textbf{Method}} 
& \multirow{2}{*}{\textbf{Data}} 
& \multicolumn{3}{c}{\textbf{Class Avg.}} 
& \multicolumn{3}{c}{\textbf{Overall Avg.}} \\
\cmidrule(lr){3-5} \cmidrule(lr){6-8}
& 
& \textbf{R} & \textbf{P} & \textbf{F1}
& \textbf{R} & \textbf{P} & \textbf{F1} \\
\midrule
SFT 
& 0\%   
& 34.6 & 31.5 & 30.0
& 30.3 & 42.3 & 35.3 \\
\midrule
\multirow{4}{*}{MWPA}
& 40\% (6K)
& 39.1 & 31.9 & 32.9
& 36.4 & 43.6 & 39.7 \\
& 60\% (9K)
& 44.6 & 32.0 & 35.4
& 45.6 & 44.8 & 45.2 \\
& 80\% (12K)
& 51.7 & 36.2 & 39.6
& 46.9 & 48.7 & 47.8 \\
& \textbf{100\% (15K)} 
& \textbf{71.8} & \textbf{38.5} & \textbf{46.9}
& \textbf{61.3} & \textbf{48.8} & \textbf{54.4} \\
\bottomrule
\end{tabular}
}
\end{minipage}
\vspace{-0.5em}
\end{table}

\begin{table}[t]
\centering
\begin{minipage}[t]{0.45\linewidth}
\centering
\captionof{table}{
\textbf{Ablation study on training stages.}
We report class-averaged and overall Recall, Precision, and F1.
}
\label{tab:ablation_stage}
\setlength{\tabcolsep}{3.4pt}
\renewcommand{\arraystretch}{1.05}
\resizebox{\linewidth}{!}{
\small
\begin{tabular}{l ccc ccc}
\toprule
\multirow{2}{*}{\textbf{Setting}}
& \multicolumn{3}{c}{\textbf{Class Avg.}}
& \multicolumn{3}{c}{\textbf{Overall Avg.}} \\
\cmidrule(lr){2-4} \cmidrule(lr){5-7}
& \textbf{R} & \textbf{P} & \textbf{F1}
& \textbf{R} & \textbf{P} & \textbf{F1} \\
\midrule
Stage1 only
& 2.6  & 11.7 & 2.7
& 2.4  & 7.9  & 3.7  \\
Stage1+2
& 34.6 & 31.5 & 30.0
& 30.3 & 42.3 & 35.3 \\
Stage1+3
& 8.2  & 27.1 & 9.7
& 6.4  & 34.0 & 10.7 \\
\rowcolor{gray!10}
Stage1+2+3
& \textbf{71.8} & \textbf{38.5} & \textbf{46.9}
& \textbf{61.3} & \textbf{48.8} & \textbf{54.4} \\
\bottomrule
\end{tabular}
}
\end{minipage}
\hfill
\begin{minipage}[t]{0.52\linewidth}
\centering
\captionof{table}{
\textbf{Ablation study on preference data construction.}
All variants start from the same Stage1+2 streaming SFT baseline.
}
\label{tab:ablation_pref}
\setlength{\tabcolsep}{2.8pt}
\renewcommand{\arraystretch}{1.05}
\resizebox{0.96\linewidth}{!}{
\small
\begin{tabular}{l ccc ccc}
\toprule
\multirow{2}{*}{\textbf{Setting}}
& \multicolumn{3}{c}{\textbf{Class Avg.}}
& \multicolumn{3}{c}{\textbf{Overall Avg.}} \\
\cmidrule(lr){2-4} \cmidrule(lr){5-7}
& \textbf{R} & \textbf{P} & \textbf{F1}
& \textbf{R} & \textbf{P} & \textbf{F1} \\
\midrule
Streaming SFT
& 34.6 & 31.5 & 30.0
& 30.3 & 42.3 & 35.3 \\
+ Macro-Window SFT
& 37.1 & 32.2 & 31.8
& 32.8 & 43.1 & 37.3 \\
+ MWPA(Format neg.)
& 33.1 & 26.6 & 29.5
& 29.6 & 39.7 & 33.9 \\
+ MWPA(On-event neg.)
& 38.5 & 30.4 & 30.2
& 33.0 & 41.7 & 36.9 \\
+ MWPA(On-policy pairs)
& 36.5 & 29.4 & 29.0
& 31.5 & 39.7 & 35.2 \\
\rowcolor{gray!10}
+ MWPA(Cross-event neg.)
& \textbf{71.8} & \textbf{38.5} & \textbf{46.9}
& \textbf{61.3} & \textbf{48.8} & \textbf{54.4} \\
\bottomrule
\end{tabular}
}
\end{minipage}
\vspace{-0.5em}
\end{table}
\textbf{Training stages.} As shown in Table~\ref{tab:ablation_stage}
Stage1 alone achieves low event recall, showing that generic streaming supervision mainly learns the low-latency generation format but provides weak soccer-specific grounding.
Adding Stage2(soccer-domain SFT), substantially improves all metrics, indicating the importance of domain adaptation for soccer terminology, commentary style, and event expressions.
Stage3 without Stage2 brings only limited gains, suggesting that macro-window preference alignment is most effective when applied to a soccer-adapted policy.
Combining Stage2 and Stage3 further improves all metrics, raising overall F1 from 35.3 to 54.4.
This confirms that macro-window alignment complements second-level SFT by injecting event-level preference feedback over causal multi-step rollouts.

\textbf{Preference-pair construction.} As shown in Table~\ref{tab:ablation_pref}
Starting from the same Stage1+2 SFT model, we compare different rejected-response strategies and a matched-data SFT control.
Macro-Window-SFT uses the same 15K chosen macro-window trajectories as MWPA but brings only modest gains, indicating that MWPA's benefit does not simply come from additional event-centered training samples.
Format-based and on-event negatives provide limited improvement, since they either lack event semantics or mainly differ in wording under the same event context.
On-policy pairs, sampled from the Stage1+2 model and ranked by Qwen3-235B, do not improve over SFT, since rollouts from the same policy rarely contain a fluent description of a wrong event type.
By contrast, cross-event counterfactual negatives achieve the largest gains by contrasting the target event narration with a fluent human commentary from a different event type within the same match, controlling style and match context while changing event semantics.
The gains of MWPA thus rely on the semantic diversity and event-type contrast provided by off-policy human trajectories.

\begin{table}[t]
\centering
\begin{minipage}[c]{0.34\linewidth}
\caption{\textbf{Results under different judge models.}
Class-averaged / overall F1 (\%) on the 11 EPL test matches, evaluated with the same protocol.}
\label{tab:judge_robustness}
\end{minipage}
\hfill
\begin{minipage}[c]{0.62\linewidth}
\centering
\resizebox{\linewidth}{!}{
\begin{tabular}{lcccccc}
\toprule
\multirow{2}{*}{\textbf{Model}}
& \multicolumn{2}{c}{\textbf{Qwen3-235B}}
& \multicolumn{2}{c}{\textbf{GPT-5.4-Mini}}
& \multicolumn{2}{c}{\textbf{Gemini-3.6-Flash}} \\
\cmidrule(lr){2-3} \cmidrule(lr){4-5} \cmidrule(lr){6-7}
& \textit{Class} & \textit{Overall}
& \textit{Class} & \textit{Overall}
& \textit{Class} & \textit{Overall} \\
\midrule
Qwen3.5~\cite{team2026qwen3_5} & 23.7 & 37.4 & 18.0 & 22.9 & 14.4 & 20.5 \\
SoccerChat~\cite{gautam2025soccerchat} & 38.1 & 48.9 & 27.3 & 34.8 & 30.4 & 36.1 \\
StreamingVLM~\cite{xu2025streamingvlm} & 29.1 & 34.2 & 23.3 & 29.6 & 21.1 & 25.7 \\
\textbf{SoccerNarrate} & \textbf{46.5} & \textbf{53.0} & \textbf{34.1} & \textbf{42.1} & \textbf{35.6} & \textbf{43.6} \\
\bottomrule
\end{tabular}
}
\end{minipage}
\vspace{-0.5em}
\end{table}

\paragraph{Robustness to the judge model.}
To check that the comparison does not hinge on a single judge, we re-evaluate the 11 EPL test matches with GPT-5.4-Mini and Gemini-3.6-Flash under the same protocol (Table~\ref{tab:judge_robustness}).
Absolute scores vary with judge calibration, but SoccerNarrate ranks first and SoccerChat second under all three judges.

\paragraph{Additional diagnostics.}
Appendix~\ref{app:add_results} provides further analyses on per-class results, latency, data scaling, robustness, stability, error sources, and evaluation settings.
These diagnostics further support the consistency of SoccerNarrate's event-grounded improvements.
\section{Conclusion}

We present SoccerNarrate for event-grounded streaming soccer commentary across data, training, and evaluation.
SoccerNarrate bridges low-latency streaming narration with event-grounded professional soccer commentary, making full-match soccer commentary both real-time and more faithful to key events.
The framework keeps second-level low-latency inference unchanged, while using time-anchored supervision and macro-window preference alignment to provide event-level feedback over causal multi-step rollouts.
We further build SoccerNarrate-Eval to measure whether generated commentary correctly mentions key soccer events under temporal constraints.
Full-match experiments show that SoccerNarrate improves event coverage and event-aligned precision over strong offline and streaming baselines.
These findings emphasize the importance of matching the temporal extent of training supervision to the events being described.
The resulting framework provides a starting point for studying how commentary content and timing interact during continuous match narration, and for exploring richer descriptions within the same streaming setting.

\paragraph{Limitations.}
SoccerNarrate targets key-event grounding rather than the factuality of all commentary content.
With a bounded streaming context, the model may still hallucinate unsupported background facts at the start of a match or after earlier context is evicted, e.g., stating a player's season record without any supporting context.
It imitates the form of professional commentary, but the factual reliability of such content still needs improvement.
Reliable commentary also depends on how individual descriptions fit the surrounding match context and remain consistent over time.
Incorporating persistent match information and evaluating longer-term narrative coherence are therefore natural directions for future work.
Further limitations are discussed in Appendix~\ref{app:limitations}.

\begin{ack}
This work is supported by the Basic Research Program of Jiangsu (No. BK20250009), the Fundamental Research Funds for the Central Universities (No. 020214380140), the Fundamental and Interdisciplinary Disciplines Breakthrough Plan of the Ministry of Education of China (No. JYB2025XDXM118), and the Nanjing University-China Mobile Communications Group Co., Ltd. Joint Institute.
The authors declare no competing interests.
\end{ack}

\begingroup
\small
\bibliography{references}
\bibliographystyle{plainnat_abbrv}
\endgroup

\appendix

\section*{Appendix Overview}

This appendix provides additional details of SoccerNarrate and further experimental results, organized as follows:

\begin{itemize}
    \item Appendix~\ref{app:add_results} reports additional experimental analyses and diagnostic results.

    \item Appendix~\ref{app:impl_details} outlines the implementation details of training and streaming inference.

    \item Appendix~\ref{app:details_about_SoccerNarrate} describes the data construction, evaluation protocol, and preference-pair construction.

    \item Appendix~\ref{app:mws_derivation} provides the derivation of Macro-Window Preference Alignment (MWPA).

    \item Appendix~\ref{app:limitations} discusses limitations and future work.
\end{itemize}
\section{Additional Results}
\label{app:add_results}
This section provides additional experimental evidence beyond the main results, including per-class breakdowns, latency analysis, data scaling, robustness checks, stability analysis, error diagnosis, and evaluation-setting studies, which together further validate the event grounding, streaming efficiency, and evaluation reliability of SoccerNarrate.
\subsection{Analysis of Per-class key events }
\label{app:per_class}
Table~\ref{tab:perclass_rpf_selected} gives the per-class Recall, Precision, and F1 breakdown on SoccerNarrate-Eval.
SoccerNarrate achieves the best class-average and overall F1, showing that the improvement in the main table is not driven by a few dominant event classes.
Compared with StreamingVLM (FT), which only adds soccer-domain second-level SFT, SoccerNarrate brings much larger gains on rare and semantically decisive events such as red cards, yellow-red cards, corners, fouls, shots, and free kicks.
This supports our claim that event-level feedback over macro-window rollouts is more effective than micro-step SFT alone.
Offline or general models, such as SoccerChat and Qwen3.5, can obtain high recall on some categories, but their lower precision indicates more temporally unsupported event mentions.
Overall, the per-class results show that SoccerNarrate provides a better balance between coverage and temporal factuality.

\begin{table*}[t]
\caption{\textbf{Per-class Performance Comparison.}
Comparison of \textbf{Recall} (Top), \textbf{Precision} (Middle), and \textbf{F1} (Bottom) for Qwen3.5, SoccerChat, StreamingVLM, StreamingVLM (FT), and our \textbf{SoccerNarrate}.
\textit{Abbreviations:}
\textbf{YC/RC}: Yellow/Red Card, \textbf{YRC}: Yellow-Red,
\textbf{Corn}: Corner, \textbf{Sub}: Substitution, \textbf{Off}: Offside,
\textbf{S-On/Off}: Shot on/off target, \textbf{Pen}: Penalty,
\textbf{DFK/IFK}: Direct/Indirect Free Kick, \textbf{BOP}: Ball out of play.
}
\label{tab:perclass_rpf_selected}
\centering
\setlength{\tabcolsep}{2.5pt}
\definecolor{algogray}{gray}{0.92}

\resizebox{\linewidth}{!}{
\begin{tabular}{c cccccccccccccc|cc}
\toprule
\multirow{2}{*}{\textbf{Model}} & \multicolumn{16}{c}{\textbf{Recall (\%)}} \\
\cmidrule(lr){2-17}
 & Goal & YC & RC & YRC & Corn & Sub & Foul & Off & S-On & S-Off & Pen & DFK & IFK & BOP & \textit{Class Avg.} & \textit{Overall Avg.} \\
\midrule
Qwen3.5~\cite{team2026qwen3_5} &
87.6 & 21.2 & 0.0 & 0.0 & 45.9 & 51.2 & 50.0 & \textbf{35.9} & 48.4 & 45.2 & 42.1 & \textbf{76.1} & 30.9 & 61.6 & 42.6 & 52.9 \\
SoccerChat~\cite{gautam2025soccerchat} &
63.9 & 56.2 & 50.0 & 0.0 & 72.3 & 49.6 & 70.0 & 29.7 & 37.3 & 20.9 & \textbf{53.8} & 64.3 & \textbf{73.7} & \textbf{73.0} & 51.1 & \textbf{63.5} \\
StreamingVLM~\cite{xu2025streamingvlm} &
80.8 & 53.9 & 0.0 & 20.0 & 43.6 & 71.1 & 39.5 & 9.3 & 33.5 & 26.1 & 21.1 & 31.7 & 15.1 & 19.8 & 33.2 & 29.4 \\
StreamingVLM (FT) &
77.6 & 63.5 & 0.0 & 20.0 & 45.5 & 74.4 & 41.1 & 14.3 & 33.1 & 28.7 & 21.1 & 32.9 & 12.3 & 20.3 & 34.6 & 30.3 \\
\rowcolor{algogray}
\textbf{SoccerNarrate} &
\textbf{96.8} & \textbf{97.8} & \textbf{100.0} & \textbf{100.0} & \textbf{82.7} & \textbf{97.2} & \textbf{87.2} & 30.4 & \textbf{51.9} & \textbf{61.5} & 31.6 & 72.0 & 48.4 & 47.3 & \textbf{71.8} & 61.3 \\
\bottomrule
\end{tabular}
}

\vspace{4pt}

\resizebox{\linewidth}{!}{
\begin{tabular}{c cccccccccccccc|cc}
\toprule
\multirow{2}{*}{\textbf{Model}} & \multicolumn{16}{c}{\textbf{Precision (\%)}} \\
\cmidrule(lr){2-17}
 & Goal & YC & RC & YRC & Corn & Sub & Foul & Off & S-On & S-Off & Pen & DFK & IFK & BOP & \textit{Class Avg.} & \textit{Overall Avg.} \\
\midrule
Qwen3.5~\cite{team2026qwen3_5} &
12.3 & 31.6 & 0.0 & 0.0 & 15.6 & 30.4 & 34.5 & 6.1 & 27.2 & 20.8 & 2.5 & 12.2 & 20.0 & 59.6 & 19.5 & 28.8 \\
SoccerChat~\cite{gautam2025soccerchat} &
\textbf{41.9} & 21.4 & \textbf{20.0} & 0.0 & 21.5 & \textbf{42.1} & 38.1 & 14.7 & 32.7 & 26.9 & \textbf{17.5} & 12.4 & 18.8 & 62.9 & 26.5 & 35.6 \\
StreamingVLM~\cite{xu2025streamingvlm} &
23.4 & 54.2 & 0.0 & 12.5 & 45.5 & 37.5 & 48.4 & 10.3 & 34.4 & 33.1 & 5.4 & 20.6 & 26.5 & 72.2 & 30.3 & 40.9 \\
StreamingVLM (FT) &
26.8 & \textbf{57.7} & 0.0 & 16.7 & 44.8 & 38.9 & 48.8 & 15.1 & 36.8 & 35.5 & 3.9 & 19.3 & 22.9 & 74.1 & 31.5 & 42.3 \\
\rowcolor{algogray}
\textbf{SoccerNarrate} &
31.4 & 41.5 & \textbf{20.0} & \textbf{45.5} & \textbf{52.9} & 37.2 & \textbf{51.0} & \textbf{27.2} & \textbf{40.6} & \textbf{45.6} & 9.5 & \textbf{21.2} & \textbf{32.3} & \textbf{82.5} & \textbf{38.5} & \textbf{48.8} \\
\bottomrule
\end{tabular}
}

\vspace{4pt}

\resizebox{\linewidth}{!}{
\begin{tabular}{c cccccccccccccc|cc}
\toprule
\multirow{2}{*}{\textbf{Model}} & \multicolumn{16}{c}{\textbf{F1 (\%)}} \\
\cmidrule(lr){2-17}
 & Goal & YC & RC & YRC & Corn & Sub & Foul & Off & S-On & S-Off & Pen & DFK & IFK & BOP & \textit{Class Avg.} & \textit{Overall Avg.} \\
\midrule
Qwen3.5~\cite{team2026qwen3_5} &
21.6 & 25.4 & 0.0 & 0.0 & 23.3 & 38.1 & 40.9 & 10.5 & 34.9 & 28.5 & 4.7 & 21.1 & 24.3 & 60.6 & 23.8 & 37.3 \\
SoccerChat~\cite{gautam2025soccerchat} &
\textbf{50.6} & 31.0 & 28.6 & 0.0 & 33.2 & 45.6 & 49.3 & 19.6 & 34.9 & 23.5 & \textbf{26.4} & 20.8 & 29.9 & \textbf{67.6} & 32.9 & 45.6 \\
StreamingVLM~\cite{xu2025streamingvlm} &
36.3 & 54.1 & 0.0 & 15.4 & 44.5 & 49.1 & 43.5 & 9.8 & 33.9 & 29.2 & 8.6 & 24.9 & 19.3 & 31.0 & 28.5 & 34.2 \\
StreamingVLM (FT) &
39.8 & \textbf{60.4} & 0.0 & 18.2 & 45.2 & 51.1 & 44.6 & 14.7 & 34.9 & 31.7 & 6.6 & 24.4 & 16.0 & 31.9 & 30.0 & 35.3 \\
\rowcolor{algogray}
\textbf{SoccerNarrate} &
47.4 & 58.3 & \textbf{33.3} & \textbf{62.5} & \textbf{64.5} & \textbf{53.8} & \textbf{64.4} & \textbf{28.7} & \textbf{45.6} & \textbf{52.4} & 14.6 & \textbf{32.8} & \textbf{38.7} & 60.1 & \textbf{46.9} & \textbf{54.4} \\
\bottomrule
\end{tabular}
}
\end{table*}

\begin{figure}
    \centering
    \includegraphics[width=0.75\linewidth]{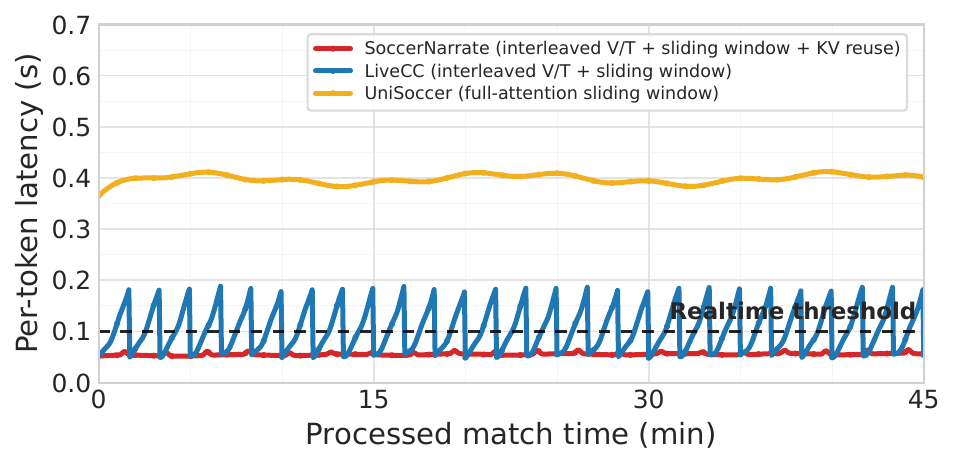}
    \caption{\textbf{Latency comparison over a 45-minute half.}
    The dashed line marks the real-time threshold of 10 tokens per second.
    SoccerNarrate maintains stable below-threshold latency with KV-cache reuse, while offline full-attention inference is much slower and LiveCC-style streaming shows periodic latency spikes.}
    \label{fig:latency_appendix}
\end{figure}

\subsection{Latency analysis}
\label{app:latency_analysis}

Figure~\ref{fig:latency_appendix} compares three inference paradigms over a 45-minute half on 1 NVIDIA H100 GPU.
We use the same real-time threshold as StreamingVLM, i.e., 10 tokens per second, corresponding to 0.1 seconds per generated token.
Offline soccer commentary models rely on full-attention sliding-window inference: each window is processed independently and the entire window context must be recomputed, resulting in consistently high per-token latency.
LiveCC-style streaming reduces the average latency by interleaving video and text tokens with a sliding window, but the lack of KV-cache reuse requires periodic window refresh or cleanup, producing sawtooth-like latency spikes and potential context discontinuities.
SoccerNarrate adopts the StreamingVLM-style streaming cache, retaining attention sinks, recent text tokens, and a short visual window while reusing KV states.
This keeps the effective context bounded and avoids redundant recomputation, leading to stable real-time latency across the full half.
Because the effective context size is bounded by the streaming cache budget, the same latency behavior is expected to extend stably to full-match and even longer streaming videos without drift.

\subsection{Effect of preference-data scaling}
\label{app:per_event_mwpa_data_scaling}
\begin{table*}[t]
\caption{\textbf{Detailed per-event results for preference-data scaling.}
Comparison of \textbf{Recall} (Top), \textbf{Precision} (Middle), and \textbf{F1} (Bottom) when using different amounts of macro-window preference data.
\textbf{Only SFT} denotes the model before preference alignment, and 100\% corresponds to the full SoccerNarrate setting with a 15s macro window.
\textit{Abbreviations:}
\textbf{YC/RC}: Yellow/Red Card, \textbf{YRC}: Yellow-Red,
\textbf{Corn}: Corner, \textbf{Sub}: Substitution, \textbf{Off}: Offside,
\textbf{S-On/Off}: Shot on/off target, \textbf{Pen}: Penalty,
\textbf{DFK/IFK}: Direct/Indirect Free Kick, \textbf{BOP}: Ball out of play.
}
\label{tab:ablation_pref_data_scaling}
\centering
\setlength{\tabcolsep}{2.3pt}
\definecolor{algogray}{gray}{0.92}

\resizebox{\linewidth}{!}{
\begin{tabular}{c c cccccccccccccc|cc}
\toprule
\multirow{2}{*}{\textbf{Model}} & \multirow{2}{*}{\textbf{Pref. Data}} & \multicolumn{16}{c}{\textbf{Recall (\%)}} \\
\cmidrule(lr){3-18}
 & & Goal & YC & RC & YRC & Corn & Sub & Foul & Off & S-On & S-Off & Pen & DFK & IFK & BOP & \textit{Class Avg.} & \textit{Overall Avg.} \\
\midrule
Only SFT &
0\% &
77.6 & 63.5 & 0.0 & 20.0 & 45.5 & 74.4 & 41.1 & 14.3 & 33.1 & 28.7 & 21.1 & 32.9 & 12.3 & 20.3 & 34.6 & 30.3 \\
+MWPA &
40\% &
82.4 & 59.5 & 0.0 & 0.0 & 60.0 & 83.0 & 48.5 & \textbf{38.6} & 49.6 & 32.5 & 0.0 & 46.3 & 24.8 & 22.4 & 39.1 & 36.4 \\
+MWPA &
60\% &
79.4 & 81.1 & 0.0 & 0.0 & 69.5 & 86.8 & 67.8 & 29.5 & 46.4 & 39.2 & 0.0 & 55.6 & 37.6 & 31.6 & 44.6 & 45.6 \\
+MWPA &
80\% &
82.4 & 62.2 & \textbf{100.0} & 0.0 & 71.6 & 90.6 & 75.5 & 29.5 & 49.6 & 39.2 & 0.0 & 51.9 & 39.1 & 31.9 & 51.7 & 46.9 \\
\rowcolor{algogray}
\textbf{+MWPA} &
\textbf{100\%} &
\textbf{96.8} & \textbf{97.8} & \textbf{100.0} & \textbf{100.0} & \textbf{82.7} & \textbf{97.2} & \textbf{87.2} & 30.4 & \textbf{51.9} & \textbf{61.5} & \textbf{31.6} & \textbf{72.0} & \textbf{48.4} & \textbf{47.3} & \textbf{71.8} & \textbf{61.3} \\
\bottomrule
\end{tabular}
}

\vspace{4pt}

\resizebox{\linewidth}{!}{
\begin{tabular}{c c cccccccccccccc|cc}
\toprule
\multirow{2}{*}{\textbf{Model}} & \multirow{2}{*}{\textbf{Pref. Data}} & \multicolumn{16}{c}{\textbf{Precision (\%)}} \\
\cmidrule(lr){3-18}
 & & Goal & YC & RC & YRC & Corn & Sub & Foul & Off & S-On & S-Off & Pen & DFK & IFK & BOP & \textit{Class Avg.} & \textit{Overall Avg.} \\
\midrule
Only SFT &
0\% &
26.8 & 57.7 & 0.0 & 16.7 & 44.8 & 38.9 & 48.8 & 15.1 & 36.8 & 35.5 & 3.9 & 19.3 & 22.9 & 74.1 & 31.5 & 42.3 \\
+MWPA &
40\% &
30.4 & 40.0 & 0.0 & 0.0 & \textbf{55.9} & 42.3 & 42.5 & \textbf{27.4} & \textbf{42.2} & 36.4 & 0.0 & \textbf{31.2} & 27.7 & 70.7 & 31.9 & 43.6 \\
+MWPA &
60\% &
25.7 & \textbf{61.2} & 0.0 & 0.0 & 43.4 & \textbf{44.7} & \textbf{51.0} & 15.9 & 39.2 & 33.3 & 0.0 & 30.3 & 31.6 & 71.3 & 32.0 & 44.8 \\
+MWPA &
80\% &
\textbf{39.4} & 51.1 & \textbf{25.0} & 0.0 & 54.8 & 39.7 & 50.6 & 20.6 & 40.0 & 44.8 & 0.0 & 25.0 & \textbf{35.1} & 80.2 & 36.2 & 48.7 \\
\rowcolor{algogray}
\textbf{+MWPA} &
\textbf{100\%} &
31.4 & 41.5 & 20.0 & \textbf{45.5} & 52.9 & 37.2 & \textbf{51.0} & 27.2 & 40.6 & \textbf{45.6} & \textbf{9.5} & 21.2 & 32.3 & \textbf{82.5} & \textbf{38.5} & \textbf{48.8} \\
\bottomrule
\end{tabular}
}

\vspace{4pt}

\resizebox{\linewidth}{!}{
\begin{tabular}{c c cccccccccccccc|cc}
\toprule
\multirow{2}{*}{\textbf{Model}} & \multirow{2}{*}{\textbf{Pref. Data}} & \multicolumn{16}{c}{\textbf{F1 (\%)}} \\
\cmidrule(lr){3-18}
 & & Goal & YC & RC & YRC & Corn & Sub & Foul & Off & S-On & S-Off & Pen & DFK & IFK & BOP & \textit{Class Avg.} & \textit{Overall Avg.} \\
\midrule
Only SFT &
0\% &
39.8 & 60.4 & 0.0 & 18.2 & 45.2 & 51.1 & 44.6 & 14.7 & 34.9 & 31.7 & 6.6 & 24.4 & 16.0 & 31.9 & 30.0 & 35.3 \\
+MWPA &
40\% &
44.4 & 47.8 & 0.0 & 0.0 & 57.9 & 56.1 & 45.3 & \textbf{32.1} & \textbf{45.6} & 34.4 & 0.0 & 37.3 & 26.2 & 34.0 & 32.9 & 39.7 \\
+MWPA &
60\% &
38.8 & \textbf{69.8} & 0.0 & 0.0 & 53.4 & \textbf{59.0} & 58.2 & 20.6 & 42.5 & 36.0 & 0.0 & \textbf{39.2} & 34.4 & 43.8 & 35.4 & 45.2 \\
+MWPA &
80\% &
\textbf{53.3} & 56.1 & \textbf{40.0} & 0.0 & 62.1 & 55.2 & 60.6 & 24.3 & 44.3 & 41.8 & 0.0 & 33.7 & 37.0 & 45.7 & 39.6 & 47.8 \\
\rowcolor{algogray}
\textbf{+MWPA} &
\textbf{100\%} &
47.4 & 58.3 & 33.3 & \textbf{62.5} & \textbf{64.5} & 53.8 & \textbf{64.4} & 28.7 & \textbf{45.6} & \textbf{52.4} & \textbf{14.6} & 32.8 & \textbf{38.7} & \textbf{60.1} & \textbf{46.9} & \textbf{54.4} \\
\bottomrule
\end{tabular}
}
\end{table*}
Table~\ref{tab:ablation_pref_data_scaling} presents per-event and averaged results for preference-data scaling, showing how the amount of macro-window preference data affects event-grounded streaming commentary.
Starting from the Only-SFT baseline, increasing the preference data generally improves both event coverage and event-aligned quality.
The gains are especially clear on averaged metrics: class-averaged Recall increases from 34.6 to 71.8, overall Recall from 30.3 to 61.3, class-averaged F1 from 30.0 to 46.9, and overall F1 from 35.3 to 54.4.
This suggests that macro-window preference alignment benefits from broader event-level supervision, rather than relying only on a small set of preference pairs.

The improvement is mainly driven by better coverage of key and long-tail events.
The SFT-only model already learns the streaming narration format and can cover frequent events such as goals, yellow cards, and substitutions to some extent, but it remains weak on rare or visually subtle categories such as red cards, yellow-red cards, penalties, and indirect free kicks.
As the preference data increases, the model becomes more likely to mention these event types under temporal constraints, leading to a large recall gain.
Precision is less monotonic at the per-class level, reflecting the usual recall--precision trade-off and the limited number of rare events.
Nevertheless, both class-averaged and overall Precision improve from the SFT baseline to the full preference setting.
These results indicate that scaling event-mismatched preference pairs helps the model learn more complete event-level commentary semantics while preserving the low-latency streaming inference protocol.

\subsection{Per-league robustness}
\providecolor{algogray}{gray}{0.9}
\begin{table*}[t]
\centering
\caption{Per-league performance comparison on SoccerNarrate-Eval. Each metric is split into class-average and overall-average columns by league. Avg. reports the corresponding full-test-set result from the main table. Test-set matches: \textbf{EPL}: 11, \textbf{LaLiga}: 10, \textbf{Bundesliga}: 6, \textbf{Serie A}: 9, \textbf{UCL}: 9 (45 total). League abbreviations: \textbf{EPL}: English Premier League; \textbf{LaLiga}: Spanish La Liga; \textbf{Bundesliga}: German Bundesliga; \textbf{Serie A}: Italian Serie A; \textbf{UCL}: UEFA Champions League.}
\label{tab:league_performance}
\resizebox{\textwidth}{!}{%
\begin{tabular}{ccccccc|cccccc}
\toprule
\multirow{3}{*}{\textbf{Model}} & \multicolumn{12}{c}{\textbf{Recall (\%)}} \\
\cmidrule(lr){2-13}
 & \multicolumn{6}{c}{\textbf{Class Avg.}} & \multicolumn{6}{c}{\textbf{Overall Avg.}} \\
\cmidrule(lr){2-7}\cmidrule(lr){8-13}
 & EPL & LaLiga & Bundesliga & Serie A & UCL & \textit{Avg.} & EPL & LaLiga & Bundesliga & Serie A & UCL & \textit{Avg.} \\
\midrule
SoccerChat~\cite{gautam2025soccerchat} & 54.3 & 58.9 & 32.0 & 52.2 & 58.4 & 51.1 & 67.1 & 68.3 & 50.7 & 61.5 & 64.0 & 63.5 \\
StreamingVLM~\cite{xu2025streamingvlm} & 33.3 & 39.0 & 30.8 & 31.4 & 34.4 & 33.2 & 29.8 & 35.9 & 22.7 & 25.9 & 30.1 & 29.4 \\
\rowcolor{algogray} SoccerNarrate & 70.0 & 69.4 & 61.5 & 72.9 & 72.0 & 71.8 & 61.1 & 62.4 & 57.8 & 62.1 & 62.1 & 61.3 \\
\bottomrule
\end{tabular}%
}
\vspace{0.45em}
\resizebox{\textwidth}{!}{%
\begin{tabular}{ccccccc|cccccc}
\toprule
\multirow{3}{*}{\textbf{Model}} & \multicolumn{12}{c}{\textbf{Precision (\%)}} \\
\cmidrule(lr){2-13}
 & \multicolumn{6}{c}{\textbf{Class Avg.}} & \multicolumn{6}{c}{\textbf{Overall Avg.}} \\
\cmidrule(lr){2-7}\cmidrule(lr){8-13}
 & EPL & LaLiga & Bundesliga & Serie A & UCL & \textit{Avg.} & EPL & LaLiga & Bundesliga & Serie A & UCL & \textit{Avg.} \\
\midrule
SoccerChat~\cite{gautam2025soccerchat} & 30.9 & 26.9 & 17.5 & 36.4 & 27.3 & 26.5 & 38.5 & 29.2 & 31.9 & 40.7 & 33.2 & 35.6 \\
StreamingVLM~\cite{xu2025streamingvlm} & 29.2 & 29.7 & 29.4 & 31.9 & 29.8 & 30.3 & 40.2 & 39.4 & 38.9 & 44.5 & 41.8 & 40.9 \\
\rowcolor{algogray} SoccerNarrate & 35.5 & 41.4 & 32.6 & 44.7 & 39.6 & 38.5 & 46.9 & 46.5 & 48.6 & 54.8 & 48.2 & 48.8 \\
\bottomrule
\end{tabular}%
}
\vspace{0.45em}
\resizebox{\textwidth}{!}{%
\begin{tabular}{ccccccc|cccccc}
\toprule
\multirow{3}{*}{\textbf{Model}} & \multicolumn{12}{c}{\textbf{F1 (\%)}} \\
\cmidrule(lr){2-13}
 & \multicolumn{6}{c}{\textbf{Class Avg.}} & \multicolumn{6}{c}{\textbf{Overall Avg.}} \\
\cmidrule(lr){2-7}\cmidrule(lr){8-13}
 & EPL & LaLiga & Bundesliga & Serie A & UCL & \textit{Avg.} & EPL & LaLiga & Bundesliga & Serie A & UCL & \textit{Avg.} \\
\midrule
SoccerChat~\cite{gautam2025soccerchat} & 38.1 & 37.0 & 21.9 & 40.0 & 38.1 & 32.9 & 48.9 & 40.9 & 39.2 & 49.0 & 43.7 & 45.6 \\
StreamingVLM~\cite{xu2025streamingvlm} & 29.1 & 30.9 & 28.1 & 26.9 & 30.3 & 28.5 & 34.2 & 37.6 & 28.6 & 32.7 & 35.0 & 34.2 \\
\rowcolor{algogray} SoccerNarrate & 46.5 & 48.9 & 43.4 & 50.6 & 51.0 & 46.9 & 53.0 & 53.3 & 52.8 & 58.2 & 54.3 & 54.4 \\
\bottomrule
\end{tabular}%
}
\end{table*}

\label{app:per_league}
Table~\ref{tab:league_performance} shows that SoccerNarrate consistently outperforms SoccerChat and StreamingVLM across all five leagues in both class-averaged and overall F1.
The improvement is especially clear in class-averaged recall, indicating that SoccerNarrate does not merely capture frequent event types but improves coverage more evenly across categories.
SoccerChat sometimes achieves competitive overall recall, but its much lower class-averaged recall and precision suggest an imbalanced tendency toward common events.
By contrast, SoccerNarrate improves recall and precision simultaneously across leagues, showing that its event-grounded gains are robust to different competitions and broadcast styles.

\subsection{Long-match stability}

\begin{figure}[t]
\centering
\includegraphics[width=0.9\linewidth]{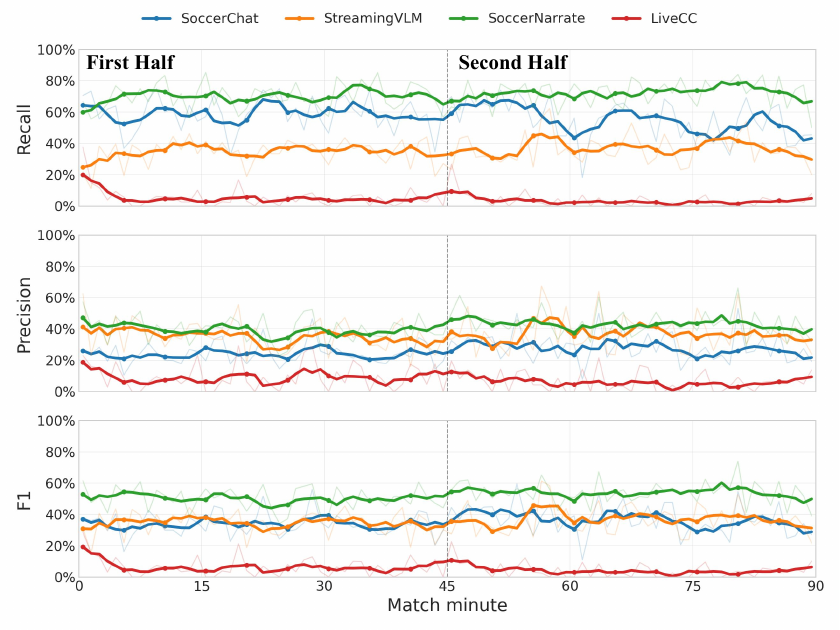}
\caption{\textbf{Long-match stability over regulation time.}
We compute event Recall, Precision, and F1 in 1-minute bins during full-match streaming inference.
Light curves show raw bin-level scores, while bold curves show smoothed trends.
The dashed line marks half time.
SoccerNarrate maintains stable event-grounded performance across both halves, without clear degradation over long-match streaming.}
\label{fig:long_match_stability}
\end{figure}

\paragraph{Long-match stability.}
Figure~\ref{fig:long_match_stability} reports event Recall, Precision, and F1 over regulation time.
We compute metrics in 1-minute bins and additionally show smoothed trends to reduce the effect of sparse event counts in individual minutes.
\textbf{(1) Stable full-match performance:} SoccerNarrate maintains strong Recall, Precision, and F1 across both halves, with no obvious degradation after half time, suggesting compatibility with long-horizon bounded-cache streaming inference.
\textbf{(2) Better balance than SoccerChat:} although SoccerChat often achieves competitive Recall, its lower Precision leads to weaker F1, indicating more temporally invalid or unsupported event mentions.
\textbf{(3) Under-coverage of StreamingVLM:} StreamingVLM remains relatively stable but clearly lower in Recall and F1, showing that second-level streaming supervision alone is insufficient for robust event coverage.
\textbf{(4) Weak soccer grounding of LiveCC:} LiveCC stays near the bottom across most bins.
\textbf{(5) Consistent gains over time:} overall, the benefits of SoccerNarrate persist throughout full-match streaming inference rather than concentrating in a few short periods.

\subsection{Robustness of the evaluation protocol}
\label{app:eval_protocol_robustness}

\providecolor{algogray}{gray}{0.9}
\begin{table*}[t]
\centering
\caption{Tolerance-window robustness on the EPL subset of SoccerNarrate-Eval. Each metric is split into class-average and overall-average columns by temporal tolerance $\delta$. The window size and stride are fixed at 30s and 15s, respectively.}
\label{tab:epl_delta_sweep}
\resizebox{\textwidth}{!}{%
\begin{tabular}{ccccccccc|cccccccc}
\toprule
\multirow{3}{*}{\textbf{Model}} & \multicolumn{16}{c}{\textbf{Recall (\%)}} \\
\cmidrule(lr){2-17}
 & \multicolumn{8}{c}{\textbf{Class Avg.}} & \multicolumn{8}{c}{\textbf{Overall Avg.}} \\
\cmidrule(lr){2-9}\cmidrule(lr){10-17}
 & 0s & 5s & 10s & 15s & 20s & 30s & 45s & 60s & 0s & 5s & 10s & 15s & 20s & 30s & 45s & 60s \\
\midrule
SoccerChat~\cite{gautam2025soccerchat} & 47.6 & 51.5 & 53.0 & 54.3 & 55.6 & 57.6 & 59.9 & 61.1 & 58.7 & 62.2 & 65.1 & 67.1 & 69.2 & 72.7 & 76.2 & 78.5 \\
StreamingVLM~\cite{xu2025streamingvlm} & 28.9 & 30.9 & 31.9 & 33.3 & 34.0 & 35.2 & 37.9 & 39.5 & 24.4 & 26.5 & 28.3 & 29.8 & 31.4 & 33.5 & 37.5 & 41.0 \\
\rowcolor{algogray} SoccerNarrate & 66.6 & 67.8 & 69.3 & 70.0 & 70.7 & 72.0 & 73.6 & 75.0 & 54.1 & 56.4 & 58.9 & 61.1 & 62.5 & 65.3 & 68.8 & 72.0 \\
\bottomrule
\end{tabular}%
}
\vspace{0.45em}
\resizebox{\textwidth}{!}{%
\begin{tabular}{ccccccccc|cccccccc}
\toprule
\multirow{3}{*}{\textbf{Model}} & \multicolumn{16}{c}{\textbf{Precision (\%)}} \\
\cmidrule(lr){2-17}
 & \multicolumn{8}{c}{\textbf{Class Avg.}} & \multicolumn{8}{c}{\textbf{Overall Avg.}} \\
\cmidrule(lr){2-9}\cmidrule(lr){10-17}
 & 0s & 5s & 10s & 15s & 20s & 30s & 45s & 60s & 0s & 5s & 10s & 15s & 20s & 30s & 45s & 60s \\
\midrule
SoccerChat~\cite{gautam2025soccerchat} & 24.4 & 27.3 & 29.5 & 30.9 & 32.7 & 35.8 & 39.6 & 41.8 & 31.0 & 34.0 & 36.7 & 38.5 & 40.8 & 44.4 & 48.3 & 51.0 \\
StreamingVLM~\cite{xu2025streamingvlm} & 23.4 & 25.8 & 27.6 & 29.2 & 30.6 & 32.6 & 35.8 & 37.8 & 31.1 & 34.4 & 37.6 & 40.2 & 42.7 & 46.5 & 52.3 & 56.9 \\
\rowcolor{algogray} SoccerNarrate & 29.7 & 31.7 & 34.0 & 35.5 & 36.7 & 39.6 & 44.0 & 45.9 & 38.3 & 41.4 & 44.5 & 46.9 & 49.0 & 52.4 & 56.7 & 59.6 \\
\bottomrule
\end{tabular}%
}
\vspace{0.45em}
\resizebox{\textwidth}{!}{%
\begin{tabular}{ccccccccc|cccccccc}
\toprule
\multirow{3}{*}{\textbf{Model}} & \multicolumn{16}{c}{\textbf{F1 (\%)}} \\
\cmidrule(lr){2-17}
 & \multicolumn{8}{c}{\textbf{Class Avg.}} & \multicolumn{8}{c}{\textbf{Overall Avg.}} \\
\cmidrule(lr){2-9}\cmidrule(lr){10-17}
 & 0s & 5s & 10s & 15s & 20s & 30s & 45s & 60s & 0s & 5s & 10s & 15s & 20s & 30s & 45s & 60s \\
\midrule
SoccerChat~\cite{gautam2025soccerchat} & 31.3 & 34.6 & 36.7 & 38.1 & 39.8 & 42.5 & 45.6 & 47.4 & 40.6 & 44.0 & 46.9 & 48.9 & 51.3 & 55.1 & 59.1 & 61.9 \\
StreamingVLM~\cite{xu2025streamingvlm} & 23.8 & 26.0 & 27.6 & 29.1 & 30.3 & 32.1 & 35.4 & 37.6 & 27.4 & 30.0 & 32.3 & 34.2 & 36.2 & 39.0 & 43.6 & 47.7 \\
\rowcolor{algogray} SoccerNarrate & 40.2 & 42.4 & 45.0 & 46.5 & 47.9 & 50.9 & 55.3 & 57.4 & 44.8 & 47.8 & 50.7 & 53.0 & 55.0 & 58.1 & 62.1 & 65.2 \\
\bottomrule
\end{tabular}%
}
\end{table*}

\providecolor{algogray}{gray}{0.9}
\begin{table*}[t]
\centering
\caption{Window-size robustness on the EPL subset of SoccerNarrate-Eval. Each metric is split into class-average and overall-average columns by window/stride setting. The temporal tolerance $\delta$ is fixed at 15s.}
\label{tab:epl_window_robustness}
\resizebox{\textwidth}{!}{%
\begin{tabular}{ccccccc|cccccc}
\toprule
\multirow{3}{*}{\textbf{Model}} & \multicolumn{12}{c}{\textbf{Recall (\%)}} \\
\cmidrule(lr){2-13}
 & \multicolumn{6}{c}{\textbf{Class Avg.}} & \multicolumn{6}{c}{\textbf{Overall Avg.}} \\
\cmidrule(lr){2-7}\cmidrule(lr){8-13}
 & 5/2.5 & 10/5 & 15/7.5 & 25/12.5 & 30/15 & 45/22.5 & 5/2.5 & 10/5 & 15/7.5 & 25/12.5 & 30/15 & 45/22.5 \\
\midrule
SoccerChat~\cite{gautam2025soccerchat} & 75.6 & 64.6 & 60.2 & 53.6 & 54.3 & 54.4 & 89.0 & 77.8 & 71.6 & 65.8 & 67.1 & 67.2 \\
StreamingVLM~\cite{xu2025streamingvlm} & 49.9 & 35.5 & 33.6 & 33.6 & 33.3 & 33.2 & 47.1 & 32.9 & 30.0 & 29.5 & 29.8 & 31.1 \\
\rowcolor{algogray} SoccerNarrate & 82.5 & 73.2 & 72.1 & 70.9 & 70.0 & 71.4 & 72.8 & 64.3 & 61.5 & 60.7 & 61.1 & 63.4 \\
\bottomrule
\end{tabular}%
}
\vspace{0.45em}
\resizebox{\textwidth}{!}{%
\begin{tabular}{ccccccc|cccccc}
\toprule
\multirow{3}{*}{\textbf{Model}} & \multicolumn{12}{c}{\textbf{Precision (\%)}} \\
\cmidrule(lr){2-13}
 & \multicolumn{6}{c}{\textbf{Class Avg.}} & \multicolumn{6}{c}{\textbf{Overall Avg.}} \\
\cmidrule(lr){2-7}\cmidrule(lr){8-13}
 & 5/2.5 & 10/5 & 15/7.5 & 25/12.5 & 30/15 & 45/22.5 & 5/2.5 & 10/5 & 15/7.5 & 25/12.5 & 30/15 & 45/22.5 \\
\midrule
SoccerChat~\cite{gautam2025soccerchat} & 3.6 & 7.4 & 11.6 & 22.9 & 30.9 & 44.2 & 3.6 & 9.1 & 16.0 & 30.9 & 38.5 & 53.7 \\
StreamingVLM~\cite{xu2025streamingvlm} & 6.6 & 12.3 & 16.6 & 24.5 & 29.2 & 41.2 & 7.8 & 15.7 & 21.5 & 32.9 & 40.2 & 58.1 \\
\rowcolor{algogray} SoccerNarrate & 7.0 & 14.1 & 20.0 & 29.4 & 35.5 & 51.3 & 8.4 & 18.4 & 25.9 & 39.1 & 46.9 & 65.2 \\
\bottomrule
\end{tabular}%
}
\vspace{0.45em}
\resizebox{\textwidth}{!}{%
\begin{tabular}{ccccccc|cccccc}
\toprule
\multirow{3}{*}{\textbf{Model}} & \multicolumn{12}{c}{\textbf{F1 (\%)}} \\
\cmidrule(lr){2-13}
 & \multicolumn{6}{c}{\textbf{Class Avg.}} & \multicolumn{6}{c}{\textbf{Overall Avg.}} \\
\cmidrule(lr){2-7}\cmidrule(lr){8-13}
 & 5/2.5 & 10/5 & 15/7.5 & 25/12.5 & 30/15 & 45/22.5 & 5/2.5 & 10/5 & 15/7.5 & 25/12.5 & 30/15 & 45/22.5 \\
\midrule
SoccerChat~\cite{gautam2025soccerchat} & 6.8 & 12.8 & 19.0 & 30.0 & 38.1 & 48.0 & 6.9 & 16.3 & 26.1 & 42.1 & 48.9 & 59.7 \\
StreamingVLM~\cite{xu2025streamingvlm} & 10.7 & 16.8 & 20.3 & 26.4 & 29.1 & 35.3 & 13.4 & 21.3 & 25.1 & 31.1 & 34.2 & 40.5 \\
\rowcolor{algogray} SoccerNarrate & 12.6 & 22.9 & 30.5 & 40.6 & 46.5 & 60.2 & 15.1 & 28.7 & 36.4 & 47.5 & 53.0 & 64.3 \\
\bottomrule
\end{tabular}%
}
\end{table*}

Tables~\ref{tab:epl_delta_sweep} and~\ref{tab:epl_window_robustness} analyze the robustness of SoccerNarrate-Eval on the EPL subset under different temporal tolerances and detection-window settings.
First, increasing the temporal tolerance $\delta$ consistently improves Recall, Precision, and F1 for all models, which is expected because more delayed or retrospective event mentions become matchable.
However, the relative ranking remains stable: SoccerNarrate consistently achieves the best class-averaged and overall F1 across all tolerance values.
This suggests that the gains are not an artifact of a particular matching threshold, but reflect more robust event-grounded generation.

Second, the window-size study shows a clear recall--precision trade-off.
Short windows such as 5/2.5s produce high Recall but very low Precision, because the dense overlapping windows create many fragmented or repeated event detections.
As the window size increases, Precision rises substantially while Recall decreases moderately, indicating that longer windows provide more complete semantic context for reliable event entailment.
Importantly, SoccerNarrate remains the strongest model across all window/stride settings, especially under the default 30/15s protocol.
This supports our choice of a 30s window with 15s stride as a practical balance: it is long enough to capture delayed human-style event descriptions, but still temporally constrained enough to penalize poorly timed, duplicated, or hallucinated event mentions.

\subsection{Error analysis}
\begin{table}[t]
\caption{\textbf{Error analysis of SoccerNarrate.}
We analyze the remaining errors under full-match streaming inference.
Precision-side errors are unmatched predicted event mentions, while recall-side errors are unrecalled ground-truth events.
Near-event attribution uses a 30s window around ground-truth events.}
\label{tab:error_analysis}
\centering
\setlength{\tabcolsep}{6pt}
\renewcommand{\arraystretch}{1.12}
\small
\begin{tabular}{llrr}
\toprule
\textbf{Side} & \textbf{Error source} & \textbf{Count} & \textbf{Share (\%)} \\
\midrule
\multirow{3}{*}{Precision}
& Near-event wrong type & 3,393 & 78.0 \\
& Delayed / retrospective mention & 175 & 4.0 \\
& No-nearby-event hallucination & 784 & 18.0 \\
\midrule
\multirow{4}{*}{Recall}
& Wrong nearby event & 1,539 & 58.9 \\
& Delayed mention & 331 & 12.7 \\
& Entity-focused drift & 745 & 28.5 \\
& Silent or generic miss & 0 & 0.0 \\
\bottomrule
\end{tabular}
\end{table}
\begin{table}[t]
\caption{\textbf{Representative adjacent-event confusions.}
These errors occur near real events and reflect common soccer event boundaries.}
\label{tab:adjacent_event_confusions}
\centering
\setlength{\tabcolsep}{7pt}
\renewcommand{\arraystretch}{1.10}
\small
\begin{tabular}{lr}
\toprule
\textbf{Adjacent event family} & \textbf{Count} \\
\midrule
Free kick / foul & 280 \\
Foul / card & 148 \\
Shot / goal & 110 \\
Corner / ball out of play & 95 \\
Free-kick subtype & 68 \\
Penalty / foul & 13 \\
\bottomrule
\end{tabular}
\end{table}
Table~\ref{tab:error_analysis} summarizes the remaining errors of SoccerNarrate under full-match streaming inference.
On the precision side, most false positives are not arbitrary hallucinations in empty match regions.
Instead, 78.0\% of false positives occur near a real ground-truth event but predict a different event type, and another 4.0\% are delayed or retrospective mentions.
Only 18.0\% are no-nearby-event hallucinations.

We use \textit{near-event wrong type} to denote predictions that occur within 30s of a real event but use a different event label.
This often reflects ambiguous event boundaries in soccer.
For example, a foul may lead to a free kick, a shot may immediately become a goal, and a ball-out-of-play moment may lead to a corner.
Thus, many false positives are better understood as fine-grained event-boundary errors rather than unsupported event invention.

On the recall side, missed events are also rarely caused by silence.
The model has no silent or purely generic misses under our attribution heuristic.
Instead, the dominant cases are wrong nearby events, delayed mentions, and entity-focused drift.
This suggests that the model usually reacts to important match regions, but may verbalize a nearby event, mention the event too late, or focus on players and local actions without explicitly stating the target event label.

Overall, the remaining gap mainly comes from exact event-type recovery under strict temporal matching.
This is especially difficult for dense and short-lived labels such as ball out of play, free kicks, and shot-related events, which often appear in rapid event chains.

\subsection{Macro-window size and in-window event density}
\label{app:window_density}
\begin{table}[t]
\caption{\textbf{Macro-window size and in-window event density.}
Cross-type events per window is the average number of annotated events of a different class inside the MWPA window. F1 values are taken from Table~\ref{tab:window_size_ablation}.}
\label{tab:window_density}
\centering
\setlength{\tabcolsep}{4pt}
\small
\begin{tabular}{lccccccccc}
\toprule
\textbf{Window $L$} & 1\,s & 5\,s & 10\,s & 15\,s & 20\,s & 25\,s & 30\,s & 35\,s & 40\,s \\
\midrule
Cross-type events / window & 0.04 & 0.21 & 0.38 & 0.55 & 0.71 & 0.87 & 1.04 & 1.22 & 1.39 \\
Class Avg.\ F1 & 30.2 & 32.3 & 42.5 & 46.9 & 45.1 & 43.3 & 39.7 & 43.1 & 42.8 \\
Overall Avg.\ F1 & 33.8 & 45.0 & 56.0 & 54.4 & 53.4 & 53.5 & 51.3 & 52.2 & 51.7 \\
\bottomrule
\end{tabular}
\end{table}

Table~\ref{tab:window_density} relates the macro-window length $L$ to the average number of annotated events of other classes inside the MWPA window.
For each target event and each $L$, we count the SoccerNet events of a different class that fall inside the corresponding window and average this count over all preference samples.
The density grows almost linearly with $L$, from 0.55 at 15\,s to 1.39 at 40\,s, while both F1 measures generally decrease after 15\,s.
Together with Fig.~\ref{fig:supervision_granularity_mismatch}, this suggests that longer windows first provide more complete event descriptions, but eventually include unrelated events that make the target-event preference signal less specific.
This is a correlation-based interpretation rather than a causal claim.
\section{Implementation Details}
\label{app:impl_details}

\paragraph{Three-stage training overview.}
We train our system in three stages: (i) supervised fine-tuning (SFT) on time-anchored streaming narration, (ii) continued SFT for stability and domain adaptation, and (iii) Macro-Window Preference
Alignment (MWPA).
Across stages, we follow a consistent optimization recipe and keep the visual backbone frozen to preserve the pretrained visual representation while adapting the language-side generation to the soccer commentary domain. All experiments were conducted on Nvidia H100 GPUs.

\paragraph{Stage 1--2: supervised training hyperparameters.}
\label{app:implementation_details}
Stages 1 and 2 share the same hyperparameter configuration.
We use AdamW with a cosine learning-rate schedule and a warmup ratio of 0.03.
The peak learning rate is set to $1\times 10^{-5}$ and the global batch size is 512.
During training, we freeze the visual backbone and update the remaining parameters in the multimodal projector and language model.
For video inputs, we uniformly sample frames at 2 FPS to balance temporal coverage and compute efficiency.

\paragraph{Stage 3: MWPA alignment hyperparameters.}
For preference alignment, we set the macro-window length to $L=15$s and use $\beta=0.1$.
We optimize the DPO objective with AdamW using a learning rate of $1\times 10^{-6}$ for 1 epoch (240 update steps).
The reference policy is the pre-alignment model obtained at the end of Stage 2 and is kept fixed throughout Stage 3, consistent with standard DPO practice.
To mitigate the ``silence bias'' common in streaming settings, we implement the trajectory-level formulation described in Appendix~\ref{app:mws_derivation}: both chosen and rejected rollouts are generated over the same macro-window horizon $L$ from an identical streaming prefix state, ensuring comparable sequence lengths and stabilizing the preference gradients.

\paragraph{Streaming inference and KV-cache management.}
Our streaming inference implementation follows the KV-cache budgeting and eviction strategy of StreamingVLM.
At each time step, we maintain a bounded cache consisting of (i) at most 512 text tokens, (ii) a 512-token attention sink, and (iii) the most recent 16 seconds of visual frames.
Inputs are represented in an interleaved video--text format, and we retain only the most recent $K$ interleaved segments that fit within the above token/frame budgets.
When the budget is exceeded, we evict the oldest segments under a FIFO-style policy while keeping the sink tokens intact, thereby preserving short-term coherence under strict memory constraints.

\paragraph{KV-cache budget (inference).}
We cap the textual context to 512 tokens and reserve 512 attention-sink tokens.
On the visual side, we retain the most recent 16 seconds of frames (sampled at 2 FPS) in the cache.
Any excess context beyond these budgets is evicted by removing the oldest interleaved segments, matching StreamingVLM's cache policy.

\begin{table}[t]
    \centering
    \caption{\textbf{Training and inference hyperparameters.}
    Stages 1--2 use identical supervised training settings. Stage 3 performs preference alignment with MWPA.}
    \label{tab:impl_hparams}
    \setlength{\tabcolsep}{6pt}
    \small
    \begin{tabular}{lccc}
        \toprule
        \textbf{Setting} & \textbf{Stage 1 (SFT)} & \textbf{Stage 2 (SFT)} & \textbf{Stage 3 (MWPA)} \\
        \midrule
        Learning rate & $1\times10^{-5}$ & $1\times10^{-5}$ & $1\times10^{-6}$ \\
        Global batch size & 512 & 512 & 64 \\
        Optimizer & AdamW & AdamW & AdamW \\
        LR schedule & Cosine & Cosine & Cosine \\
        Warmup ratio & 0.03 & 0.03 & 0.03 \\
        Train epochs & 1 & 1 & 1 \\
        Frozen modules & Visual backbone & Visual backbone & Visual backbone \\
        Frame sampling & 2 FPS & 2 FPS & 2 FPS \\
        \midrule
        Macro-window $L$ & -- & -- & 15\,s \\
        DPO $\beta$ & -- & -- & 0.1 \\
        Reference policy & -- & -- & Stage-2 model (fixed) \\
        \bottomrule
    \end{tabular}
\end{table}

\section{Details about SoccerNarrate}
\label{app:details_about_SoccerNarrate}
\subsection{Human verification}
\label{app:data_validation}
We verify both the curation quality and the reliability of the event-entailment judge.
These human checks were conducted internally by the authors and research-group members on existing commentary; no external study participants were recruited and no crowdsourcing platform was used.
On 30 sampled full matches, spelling errors are negligible after cleaning, and entity calibration reduces the player/team name error rate from 28.4\% to 0.1\%.
As shown in Table~\ref{tab:human_validation_qwen_judge}, Qwen3-235B achieves 98.8\% overall accuracy and 97.6\% class-average accuracy on 1,843 human-labeled ASR segments.
Since SoccerNarrate-Eval is applied to model-generated commentary, we further sample 1,200 generated event-entailment queries from StreamingVLM~(FT), Matched-SFT, and SoccerNarrate.
Qwen3-235B achieves 94.7\% accuracy against human labels on these model outputs, suggesting that it remains reliable on fragmented and potentially noisy generated commentary.

\begin{table}[t]
\caption{\textbf{Human validation of Qwen3-235B event judgments.}
We compare Qwen3-235B judgments with human labels on both cleaned ASR segments and model-generated commentary.}
\label{tab:human_validation_qwen_judge}
\centering
\setlength{\tabcolsep}{6pt}
\renewcommand{\arraystretch}{1.12}
\small
\begin{tabular}{llcc}
\toprule
\textbf{Source} & \textbf{Judgment Task} & \textbf{\#Samples} & \textbf{Acc. (\%)} \\
\midrule
Cleaned ASR segments 
& Event type classification 
& 1,843 
& 98.8 / 97.6$^\dagger$ \\
GT-centered model outputs 
& Entails the GT event 
& 600 
& 94.3 \\
Prediction-centered model outputs 
& Valid predicted event mention 
& 600 
& 95.0 \\
\midrule
Model outputs overall 
& Event entailment 
& 1,200 
& 94.7 \\
\bottomrule
\end{tabular}
\vspace{2pt}
\caption*{\footnotesize $^\dagger$Overall / class-average accuracy.}
\end{table}

\paragraph{Class-balanced validation.}
To examine whether the judge is biased toward particular event classes, we further sample exactly 50 event-centered commentary windows per class and compare Qwen3-235B with human labels under the same 30-second entailment protocol (Table~\ref{tab:human_validation_classwise}).
The macro-averaged agreement is 90.4\%.
Agreement is near ceiling for explicitly verbalized events such as goals, penalties, cards, offsides, and substitutions, while most disagreements occur for linguistically ambiguous classes such as indirect free kicks, ball out of play, and corners, where commentary often describes an event chain or its consequence without naming the fine-grained label.

\begin{table}[t]
\caption{\textbf{Class-balanced human validation of the Qwen3-235B judge.}
Agreement (\%) between Qwen3-235B and human labels on 50 event-centered commentary windows per class.}
\label{tab:human_validation_classwise}
\centering
\setlength{\tabcolsep}{6pt}
\renewcommand{\arraystretch}{1.05}
\small
\begin{tabular}{lclc}
\toprule
\textbf{Event} & \textbf{Agr.} & \textbf{Event} & \textbf{Agr.} \\
\midrule
Goal & 100.0 & Corner & 80.0 \\
Penalty & 100.0 & Substitution & 96.0 \\
Red card & 96.0 & Foul & 86.0 \\
Yellow$\rightarrow$red card & 100.0 & Offside & 100.0 \\
Yellow card & 96.0 & Direct free-kick & 84.0 \\
Shots on target & 100.0 & Indirect free-kick & 66.0 \\
Shots off target & 86.0 & Ball out of play & 76.0 \\
\midrule
\multicolumn{4}{c}{\textbf{Macro average: 90.4}} \\
\bottomrule
\end{tabular}
\end{table}

\subsection{Data Source and Time-Anchored Streaming Pipeline}
SoccerNarrate is built from the official SoccerNet broadcast collection, starting with 500 full-match videos (about 90 minutes each). We remove 64 matches without commentary, leaving 436 matches for the dataset. For each match, we extract and denoise the audio to reduce background noise, making speech easier to align. The cleaned audio is transcribed with WhisperX-large v3 for word-level timestamps; non-English commentary is automatically translated to English with timestamps preserved.
We then use speaker diarization to tag each segment, keeping up to two main commentators per match and filtering out others (like interviews or crowd noise). Adjacent utterances are merged into segments of at least 3 seconds, allowing short pauses (up to 10 seconds) for natural flow.
To further clean the transcripts, we use a large language model (Qwen3-235B) to fix grammar and minor errors without changing the original meaning. For entity grounding, we extract player names from each segment and match them to official rosters using phoneme-based matching, correcting noisy spellings. Finally, another LLM-based filter removes low-information segments (such as fillers or off-topic speech), cutting about 20\% of the total commentary duration. The final dataset covers 74.46\% of total video length. Following SoccerNet’s split, the training set has 391 matches and the test set 45 matches. For training, all commentary is split into 16–32 second windows, allowing up to 10 seconds of internal silence per segment.

\begin{figure*}[!ht]
  \centering
  \includegraphics[width=1\textwidth]{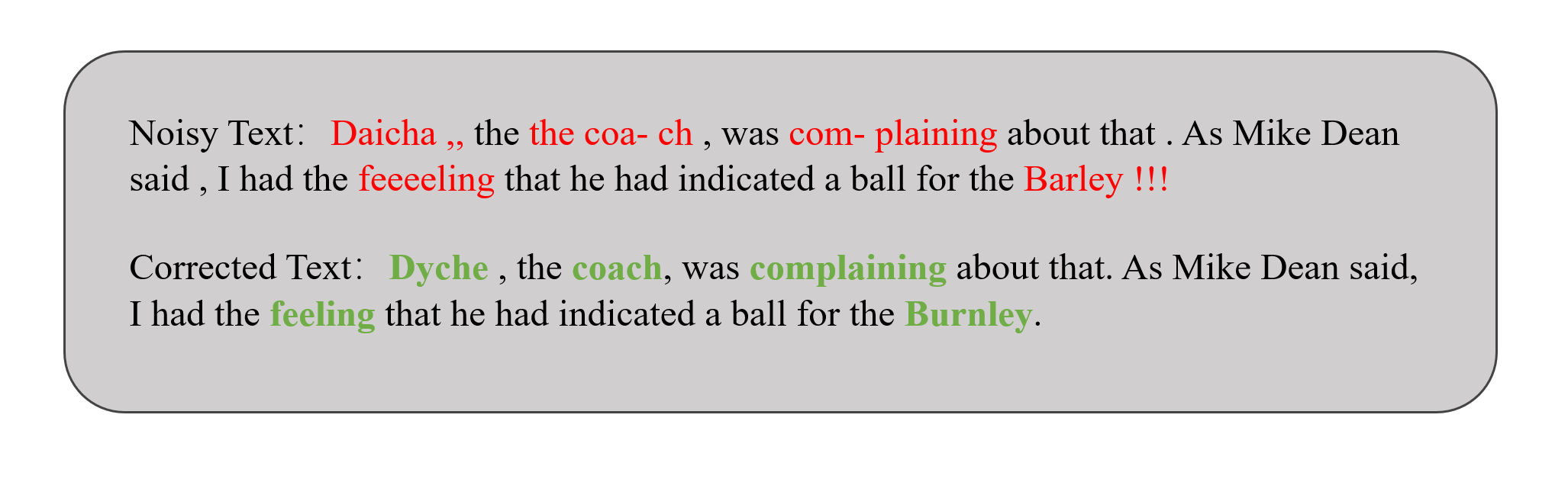}
  \caption{\textbf{Qualitative Results of Entity Correction.}The figure demonstrates the pipeline's capability to handle compound noise patterns. The Noisy Text (top) contains multiple error types highlighted in red, including phonetic entity errors (e.g., "Daicha"), word fragmentation (e.g., "coa- ch"), token repetition, and irregular punctuation. The Corrected Text (bottom) displays the final output with all restored and normalized segments highlighted in green, verifying the model's robustness in simultaneously performing de-duplication, de-fragmentation, and entity correction.}
  \label{fig:Corrected}
\end{figure*}

\subsection{Data Statistics and Analysis}
\begin{figure*}[t]
    \centering
    \includegraphics[width=\linewidth]{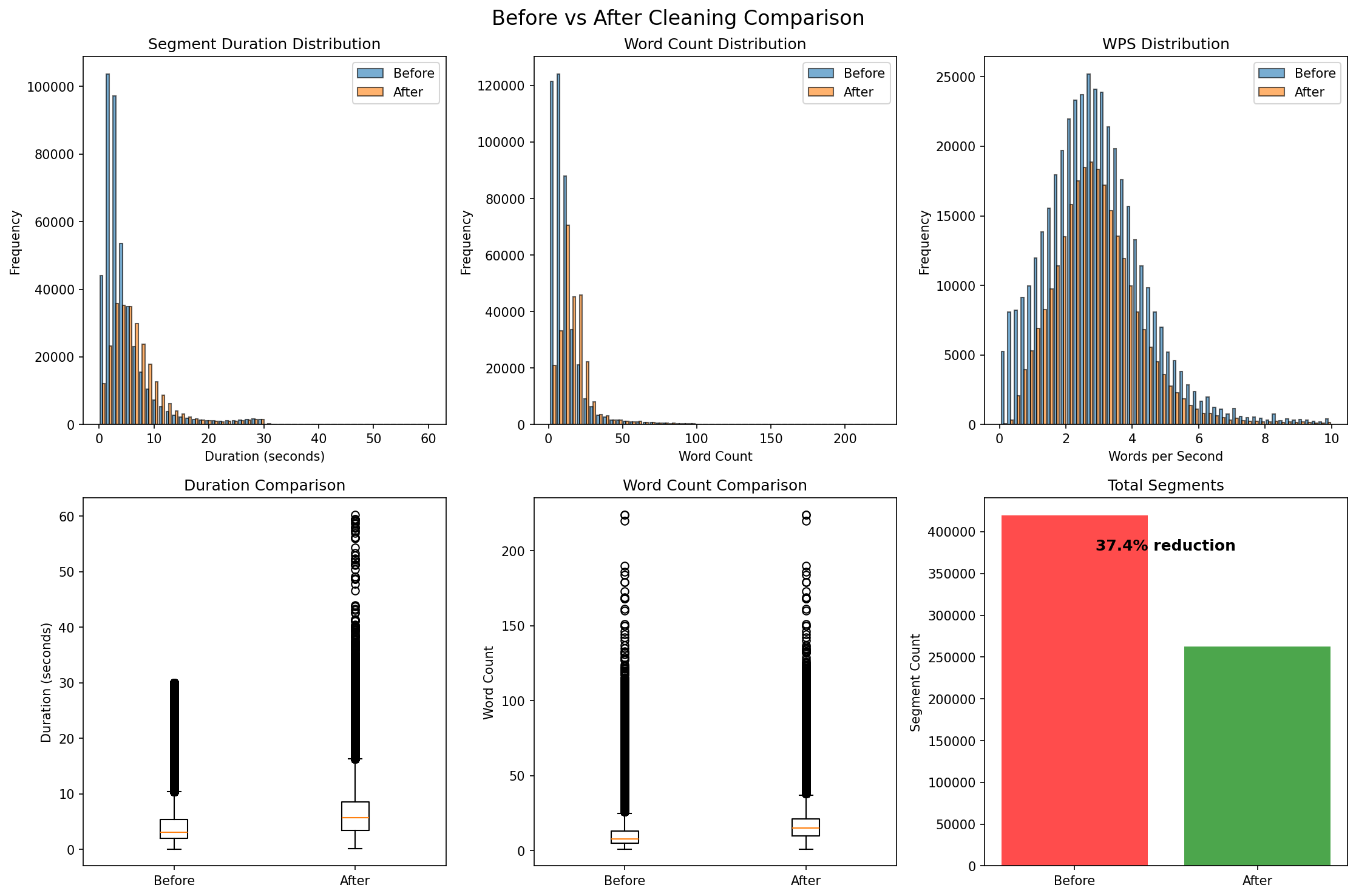}
    \caption{\textbf{Before vs. after cleaning statistics for SoccerNarrate.}
    We compare segment duration, word count, and words-per-second (WPS) distributions, together with boxplot summaries and the total segment count.
    The cleaning pipeline shifts the dataset toward longer, more coherent narration segments while largely preserving speaking-rate statistics, and reduces the number of segments by 37.4\% through quality gating.}
    \label{fig:data_stats}
\end{figure*}

\cref{fig:data_stats} summarizes how our quality-gated cleaning pipeline reshapes the segment-level statistics. 
Before cleaning, the narration stream is dominated by extremely short utterances, yielding a heavy mass of sub-3s segments and correspondingly low word counts; such fragments are often low-information fillers and provide weak supervision for event-centric learning. 
After cleaning, the duration and word-count distributions shift toward longer and more coherent segments, consistent with our merging constraints (minimum segment length and bounded internal silence) and the removal of diarization/ASR artifacts. 
Importantly, the words-per-second (WPS) distribution remains largely stable and becomes slightly more concentrated, indicating that the pipeline primarily improves temporal grouping and textual fidelity rather than inflating content density via aggressive rewriting. 
Overall, the total number of segments decreases by 37.4\%, reflecting substantial pruning of redundant or low-quality commentary while preserving temporally anchored, information-rich narration segments that are better suited for streaming supervision and preference alignment.

\subsection{SoccerNarrate-Eval: Event Entailment under Temporal Matching}
\label{app:soccernarrate_eval}

\paragraph{Core idea: detecting event mentions in generated commentary.}
SoccerNarrate-Eval evaluates whether a streaming commentator mentions key soccer events correctly and on time.
Unlike captioning metrics or pairwise win rates, our protocol converts free-form generated commentary into event mentions and then matches these mentions to ground-truth SoccerNet events under temporal constraints.
This separates two aspects of evaluation: \emph{what} event is mentioned and \emph{when} it is mentioned.
A generated event mention is considered correct only if it entails the same event type as a ground-truth event and is temporally close enough to that event.

\paragraph{1) Sliding-window event detection.}
For each match half, we collect the model-generated commentary produced at one-second streaming steps, together with its timestamps.
We normalize the raw text by removing formatting artifacts, trimming duplicated overlaps, and standardizing whitespace.
We then slide a temporal window over the generated commentary with window size $W=30\,\mathrm{s}$ and stride $S=15\,\mathrm{s}$.
Each window is denoted as $\hat{w}=[\hat{s},\hat{e})$.
For every window, we concatenate all generated text whose timestamps fall inside $\hat{w}$ into a single \texttt{COMMENTARY\_TEXT}.
Any event mention detected from this window is localized by the whole window interval rather than by a single timestamp.

\paragraph{2) Entailment-based event extraction.}
We use Qwen3-235B as a conservative event-entailment detector.
For each generated-commentary window $\hat{w}$, the judge determines whether the window text entails any of the 14 evaluated event categories.
The evaluated categories are Goal, Yellow Card, Red Card, Yellow-Red Card, Corner, Substitution, Foul, Offside, Shot on Target, Shot off Target, Penalty, Direct Free Kick, Indirect Free Kick, and Ball Out of Play.
We exclude Kick-off, Throw-in, and Clearance because they are frequent routine actions and are rarely verbalized explicitly in live commentary.
If the judge detects an event category $\hat{c}$ in window $\hat{w}$, we create a predicted event mention $(\hat{w},\hat{c})$ that keeps the interval of its source window.
If multiple event categories are entailed in the same window, each detected category is treated as a separate predicted mention.

\paragraph{3) Temporal one-to-one matching.}
Let $\mathcal{G}=\{(t_i,c_i)\}$ denote the set of ground-truth SoccerNet event annotations over the 14 evaluated classes, and let $\mathcal{P}=\{(\hat{w}_j,\hat{c}_j)\}$ denote the set of predicted event mentions extracted from generated commentary, where $\hat{s}_j$ and $\hat{e}_j$ are the start and end of window $\hat{w}_j$.
We measure the temporal distance from a ground-truth timestamp $t$ to a window interval as
\begin{equation}
d\big(t,[\hat{s},\hat{e}]\big)=\max\big(\hat{s}-t,\;0,\;t-\hat{e}\big),
\end{equation}
which is zero when $t\in[\hat{s},\hat{e}]$ and otherwise equals the distance to the nearest window boundary.
A predicted mention $(\hat{w}_j,\hat{c}_j)$ can match a ground-truth event $(t_i,c_i)$ only if the event type is the same and this distance is within a tolerance $\delta$:
\begin{equation}
\hat{c}_j=c_i
\quad\text{and}\quad
d\big(t_i,[\hat{s}_j,\hat{e}_j]\big)\le \delta ,
\end{equation}
where we use $\delta=15\,\mathrm{s}$ by default.
Equivalently, the ground-truth event must occur within ${[\hat{s}_j-\delta,\,\hat{e}_j+\delta]}$.
We enforce one-to-one matching so that each ground-truth event can be matched to at most one predicted mention, and each predicted mention can be matched to at most one ground-truth event.
When multiple valid pairs are possible, we greedily select the pair with the smallest distance $d$ and remove the matched prediction and ground-truth event from further matching.
This prevents repeated mentions of the same event from being counted as multiple correct hits.

\paragraph{4) Metrics.}
Let $\mathcal{M}$ be the final set of one-to-one matched prediction--ground-truth pairs.
We compute event-level Recall, Precision, and F1 as
\begin{equation}
\mathrm{Recall}=\frac{|\mathcal{M}|}{|\mathcal{G}|},
\qquad
\mathrm{Precision}=\frac{|\mathcal{M}|}{|\mathcal{P}|},
\qquad
\mathrm{F1}=\frac{2\cdot \mathrm{Recall}\cdot \mathrm{Precision}}
{\mathrm{Recall}+\mathrm{Precision}}.
\end{equation}
Recall measures the fraction of ground-truth events that are correctly mentioned by the generated commentary.
Precision measures the fraction of predicted event mentions that are both semantically correct and temporally aligned.
Thus, hallucinated, duplicated, wrong-type, or poorly timed event mentions are counted as false positives, while ground-truth events without a matched mention are counted as false negatives.
We report both class-averaged metrics, which average over event classes, and overall metrics, which aggregate all event instances before computing the scores.

\paragraph{LLM-judge prompts.}
We employ Qwen3-235B as the entailment detector.
All prompts enforce conservative decisions and strict JSON outputs.
Malformed outputs or predictions without explicit textual evidence are downgraded to \texttt{not\_happened}.

\newcommand{\promptbox}[1]{%
\vspace{3pt}
\noindent\fcolorbox{black!25}{black!5}{%
\begin{minipage}{0.98\linewidth}
\small\ttfamily #1
\end{minipage}}%
\vspace{3pt}
}

\promptbox{
\textbf{System prompt} \\
You are a careful soccer (football) commentary analysis assistant. \\
\\
You will be given a short window of match commentary text. \\
\\
Rules: \\
- Use ONLY the given text. \\
- Be conservative: if the text is unclear or ambiguous, choose not\_happened. \\
- Only mark an event as happened when the commentary explicitly or strongly implies that event. \\
\\
Return JSON only and follow the requested schema exactly.
}

\promptbox{
\textbf{Event entailment user prompt} \\
Below is a window of live soccer commentary text: \\
\{\{COMMENTARY\_TEXT\}\} \\
\\
Candidate event labels: \\
Goal; Yellow Card; Red Card; Yellow-Red Card; Corner; Substitution; Foul; Offside; Shot on Target; Shot off Target; Penalty; Direct Free Kick; Indirect Free Kick; Ball Out of Play. \\
\\
Question: \\
Which candidate event labels are entailed by this commentary window? \\
Only include labels that are explicitly mentioned or strongly implied by the text. \\
If no event is entailed, return an empty list. \\
\\
Return strict JSON only: \\
\{"events":[\{"label":"<event label>","evidence":"<short quote or rationale>"\}]\}
}

\subsection{Order Sensitivity of Pairwise Win-rate}
\label{app:winrate_bias}

Pairwise LLM-as-a-judge win-rate is commonly used to evaluate open-ended commentary, since different systems may describe the same event with diverse surface forms.
However, we find that this metric is sensitive to prompt order and does not always reliably isolate key-event correctness.
To examine this issue, we construct a controlled counterfactual event-swap test.

We sample 1,100 key-event-centered windows from the SoccerNet test split.
For each event, we use the 30s ASR commentary around the ground-truth event as the reference.
The event-correct candidate is the StreamingVLM output from the same event window, obtained after running full-match inference.
The counterfactual candidate is replaced by a same-model output from another window, either corresponding to a different event type, such as replacing a goal description with a yellow-card description, or containing background/generic commentary without a key event.
Since both candidates are generated by the same model, this setup largely controls for fluency, length, and commentary style, while changing whether the candidate matches the target event.

Table~\ref{tab:winrate_order_bias} shows the results.
When the prompt order is \texttt{GT, Correct, Counterfactual}, the event-correct candidate wins 78.0\% of the time against event-mismatched key-event commentary, and 68.0\% against background/generic commentary.
However, when we keep the reference and candidate contents unchanged but swap the candidate order to \texttt{GT, Counterfactual, Correct}, the win rate of the event-correct candidate drops to 29.0\% and 23.0\%, respectively.
This indicates that pairwise win-rate can exhibit strong positional bias and may fail to consistently prefer the event-correct candidate.
Therefore, we treat win-rate as a complementary holistic metric rather than an independent measure of key-event faithfulness, and use temporally constrained event entailment as the main evaluation protocol.
\begin{table}[t]
\centering
\small
\caption{\textbf{Order sensitivity of pairwise win-rate.}
We report the win rate of the event-correct candidate in a controlled event-swap test on 1,100 event-centered SoccerNet test windows.}
\label{tab:winrate_order_bias}
\setlength{\tabcolsep}{8pt}
\renewcommand{\arraystretch}{1.15}
\begin{tabular}{lcc}
\toprule
\textbf{Counterfactual type} 
& \textbf{Correct first} 
& \textbf{Correct second} \\
\midrule
Event-mismatched key event & 78.0 & 29.0 \\
Background / generic commentary & 68.0 & 23.0 \\
\bottomrule
\end{tabular}
\end{table}

\subsection{Human Event Recall and Precision in Streaming Soccer Commentary}
\label{sec:human_upper_bound}

\paragraph{Motivation.}
A key point in streaming soccer commentary is that human commentators do not always mention every SoccerNet-labeled event. Sometimes they stay silent, delay their comments, or focus on broader context instead of listing every small event. So, if we evaluate models only based on SoccerNet labels, we may unfairly penalize them for missing events that humans also skip. To set a fair baseline, we measure and report how often human commentators recall these events and, later, their precision using the same evaluation protocol.

\begin{figure*}[t]
    \centering

    \begin{minipage}[c]{0.55\linewidth}
        \centering
        \vspace*{\fill}
        \includegraphics[width=0.95\linewidth]{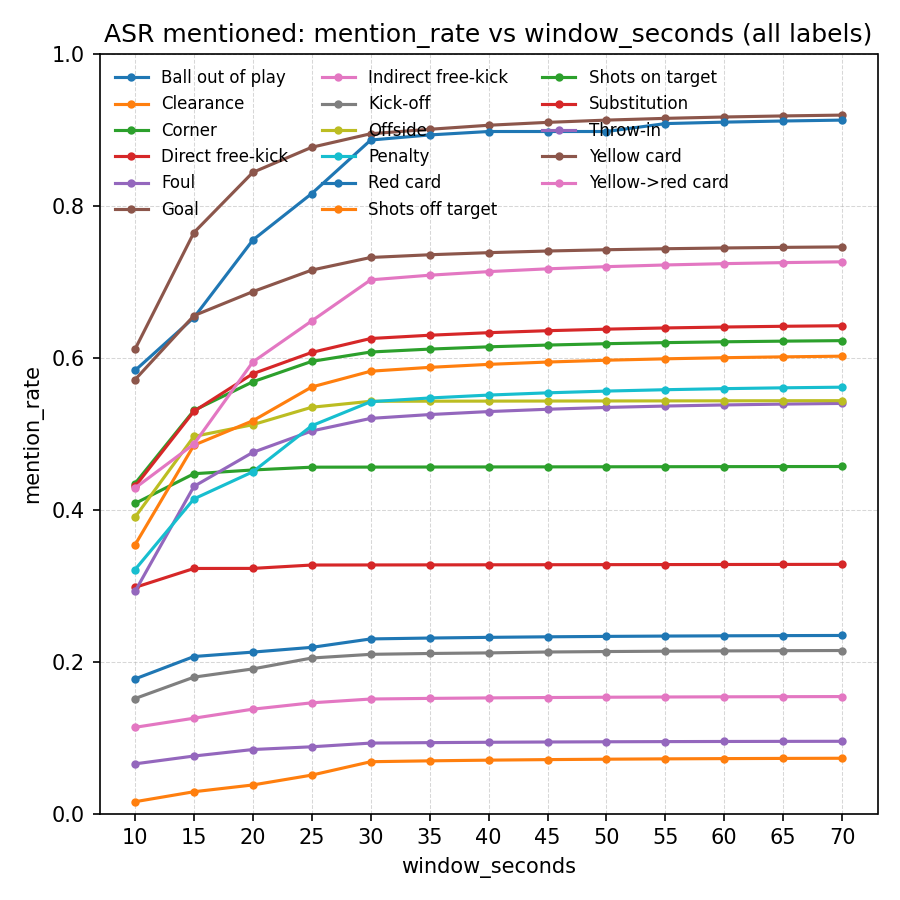}
        \par\vspace{2pt}
        \small\textbf{(Left)} Human recall (mention rate) vs.\ entailment window length.
        \vspace*{\fill}
    \end{minipage}\hfill
    \begin{minipage}[c]{0.45\linewidth}
        \centering
        \setlength{\tabcolsep}{5pt}
        \footnotesize
        \begin{tabular}{lccc}
            \toprule
            \textbf{Event Class} & \textbf{Recall (\%)} & \textbf{Total} \\
            \midrule
            Goal & 89.5 & 1451 \\
            Red card & 88.7 & 49 \\
            Yellow card & 73.2 & 1591 \\
            Yellow$\rightarrow$red card & 70.3 & 37 \\
            Substitution & 62.5 & 1969 \\
            Shots on target & 60.8 & 4988 \\
            Shots off target & 58.2 & 4495 \\
            Offside & 54.3 & 1638 \\
            Penalty & 54.2 & 142 \\
            Foul & 52.0 & 9457 \\
            Corner & 45.5 & 3551 \\
            Direct free-kick & 31.8 & 1519 \\
            Ball out of play & 22.9 & 25056 \\
            Kick-off & 21.0 & 569 \\
            Indirect free-kick & 15.1 & 4702 \\
            \midrule
            Throw-in & 9.3 & 10774 \\
            Clearance & 6.8 & 2921 \\
            \bottomrule
        \end{tabular}
        \par\vspace{2pt}
        \small\textbf{(Right)} Human recall by class (30\,s window).
    \end{minipage}

    \caption{\textbf{Human mention behavior under entailment-based evaluation.}
    Left: recall increases with entailment window length, motivating a 30\,s default as a trade-off between delayed mentions and cross-event contamination.
    Right: per-class human recall at 30\,s.}
    \label{fig:human_recall_combo}
\end{figure*}

\paragraph{Human event recall as \emph{mention rate}.}
We operationalize \emph{human recall} as whether the human commentary \emph{mentions} a labeled event within a bounded temporal neighborhood, using the same entailment-based judge as SoccerNarrate-Eval (Sec.~\ref{app:soccernarrate_eval}).
For each GT event instance $(\ell,t)$, we extract the human commentary within a window around $t$ and ask the LLM judge whether the text entails that event $\ell$ happened in that window.
Table~\ref{fig:human_recall_combo} reports the resulting per-class human recall (mention rate) together with the total number of GT instances.

\paragraph{Window-length trade-off for entailment-based recall.}
Figure~\ref{fig:human_recall_combo} analyzes how the measured human recall varies with the entailment window size.
As the window length increases, recall monotonically improves for nearly all event types, since longer windows are more likely to include explicit mentions, delayed reactions, or short retrospective descriptions.
However, overly large windows introduce substantial \emph{semantic confounds}: they aggregate multiple discourse topics, may overlap with subsequent key events, and increase background narration that can dilute the event-specific evidence available to the judge.
In practice, this increases the risk of noisy or ambiguous entailment decisions and weakens the desired notion of \emph{time-anchored mention}.
We therefore adopt a \textbf{30\,s} window as the default (with the same tolerance $\delta$ in Sec.~\ref{app:soccernarrate_eval}), striking a practical trade-off between capturing delayed human mentions and limiting cross-event contamination.

\paragraph{Key observation: humans do \emph{not} narrate every labeled event.}
Across window settings, a pronounced long-tail persists: high-salience events (e.g., Goal, cards, penalties, substitutions) are frequently mentioned, whereas high-frequency background actions are often omitted.
\textit{Throw-in} and \textit{Clearance} exhibit consistently low recall and are comparatively low in narrative value for broadcast commentary.
We also exclude \textit{Kick-off}, which mainly denotes a predictable match-phase boundary and can often be inferred from the clock or match state, rather than requiring visually grounded event recognition.
Following this observation, we focus our main per-class reporting on \textbf{14 event types}, excluding \textit{Kick-off}, \textit{Throw-in}, and \textit{Clearance}.
This choice better matches the communicative objective of live commentary and avoids over-emphasizing low-value or template-like classes.

\paragraph{Human precision.}
Table~\ref{tab:human_precision} shows that human precision under strict temporal matching is lower than expected (e.g., 13.5\% for Penalty).
It is important to clarify that our metric measures \textbf{strict temporal alignment}, not just factual correctness.
Low precision here reflects the human stylistic tendency to anticipate events early or review them late, causing their narration to drift outside the strict event window.
This indicates that humans prioritize storytelling flow over precise timestamp synchronization.
Therefore, precision should not be viewed in isolation. 
A comprehensive evaluation must combine Strict Precision (how timely and aligned the narration is) with Recall (how many events are covered) to fully understand the trade-off between grounding and coverage.


\begin{table}[t]
\centering
\small
\begin{tabular}{l c}
\hline
\textbf{Label} & \textbf{Precision (\%)} \\
\hline
Ball out of play      & 81.6 \\
Corner                & 56.3 \\
Direct free-kick       & 29.9 \\
Foul                  & 50.0 \\
Goal                  & 24.9 \\
Indirect free-kick     & 34.3 \\
Offside               & 45.9 \\
Penalty               & 13.5 \\
Red card              & 68.2 \\
Shots off target      & 48.4 \\
Shots on target       & 43.3 \\
Substitution          & 35.7 \\
Yellow card           & 58.3 \\
Yellow$\rightarrow$red card & 10.5 \\
\hline
\textbf{Class-average} & \textbf{38.5} \\
\textbf{Overall-average} & \textbf{48.0} \\
\hline
\end{tabular}
\caption{Human precision under the same event-entailment and time-matching protocol.}
\label{tab:human_precision}
\end{table}

\begin{figure}
    \centering
    \includegraphics[width=\linewidth]{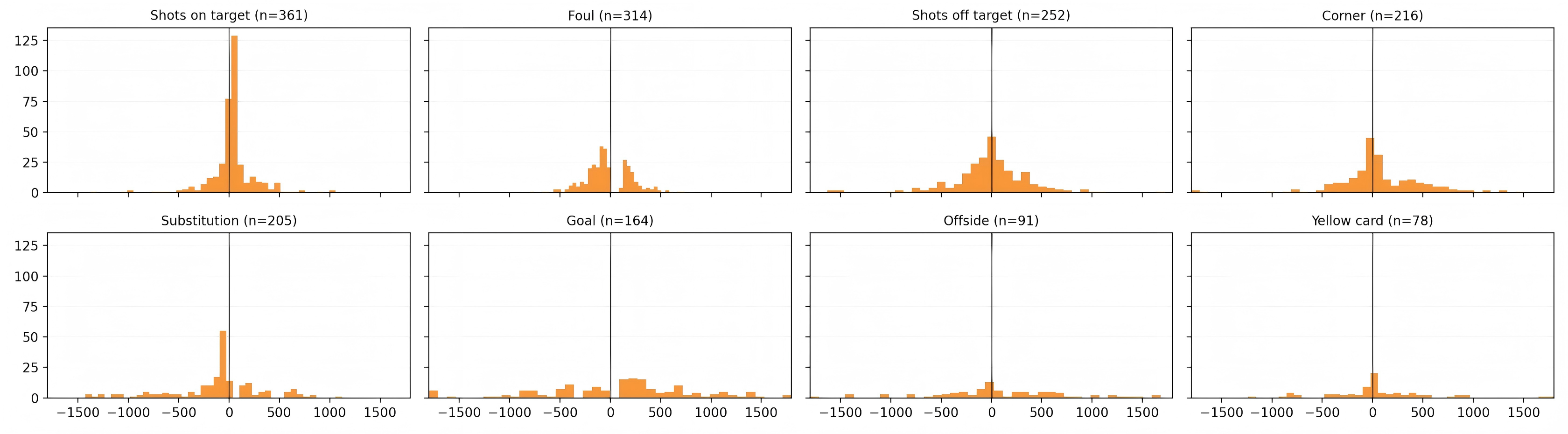}
    \caption{\textbf{Temporal distribution of commentary sentences relative to the official event start.} The x-axis represents the time offset in seconds (where the dashed line at 0 marks the event start), and the y-axis indicates the count of commentary sentences for different event types.}
    \label{fig:otm}
\end{figure}

\subsection{SoccerNarrate preference pair}
\label{sec:pref_pairs}

\paragraph{Source and principle.}
All preference pairs are constructed exclusively from the \textbf{SoccerNarrate} training split, using SoccerNet-Labels as the event anchor. 
SoccerNet-Labels provides precise timestamps for 17 event types per match, enabling event-conditioned supervision with strict temporal grounding.
For each labeled event instance at time $t$, we form a macro-window commentary segment by extracting the human narration within a fixed interval $[t-5\text{s},\, t+25\text{s}]$. 
We then apply an entailment-based filtering step using Qwen3-235B to verify whether the extracted narration segment semantically entails the target event (details in the next subsection). 
Only segments that entail the target event are retained, and the corresponding human narration is treated as the \textbf{chosen} response for that annotated target event instance.

\paragraph{Counterfactual hard negatives with matched style and length.}
For each chosen segment anchored at event $(e,t)$, we construct the \textbf{rejected} response by selecting a event-mismatched instance $(e',t')$ from the same match, where $e' \neq e$, and extracting its narration using the identical temporal window $[t'-5\text{s},\, t'+25\text{s}]$. 
This cross-event pairing yields a counterfactual negative: the rejected narration is fluent and stylistically consistent (typically produced by the same commentator(s) within the same broadcast), yet it is non-aligned with the anchor event by construction. 
Crucially, using the same window length keeps the total response length closely matched between chosen and rejected, preventing trivial preferences driven by verbosity rather than correctness.

\paragraph{Comparable prefixes under streaming constraints.}
To satisfy the DPO assumption of a shared conditioning context while respecting streaming inference constraints, we enforce a shared prefix state for each pair.
Specifically, both chosen and rejected rollouts are conditioned on the same streaming prefix consisting of (i) the most recent 16\,s of video context (to approximate the bounded KV-cache in online decoding) and (ii) the most recent 512 text tokens from the running commentary history. 
This design ensures that the preference signal reflects event alignment within the macro window rather than differences in accessible history, and makes the preference objective compatible with low-latency streaming policies.

\paragraph{Scale and event-balanced sampling.}
In total, we construct $\sim$15K preference pairs. 
Because SoccerNet event frequencies are highly imbalanced and different event types carry different narrative salience, we stratify event types into three tiers and allocate pair budgets accordingly.
We use tier-level mixture weights to prioritize high-value, event-defining moments (T1), while still retaining moderate coverage of secondary actions (T2) and a small fraction of high-frequency background events (T3). 
Within each tier, we further specify a fixed anchor-event sampling distribution to determine the final per-class pair counts and proportions.

\begin{table}[t]
    \centering
    \caption{\textbf{Event tiers and anchor sampling priors for preference-pair construction.}
    We group the 17 SoccerNet event types into three narrative tiers and assign tier-level mixture weights.
    Within tiers, we use a fixed anchor-event probability prior to allocate the 15K preference-pair budget across classes.}
    \label{tab:prefpair_tiers_probs}
    \setlength{\tabcolsep}{5pt}
    \small
    \begin{tabular}{ccc}
        \hline
        \textbf{Event Type} & \textbf{Tier} & \textbf{Anchor Prob.} \\
        \hline
        Goal & T1 & 0.136 \\
        Penalty & T1 & 0.022 \\
        Red card & T1 & 0.016 \\
        Yellow$\rightarrow$red card & T1 & 0.013 \\
        Yellow card & T1 & 0.084 \\
        Shots on target & T1 & 0.106 \\
        Shots off target & T1 & 0.089 \\
        Corner & T1 & 0.099 \\
        Substitution & T1 & 0.086 \\
        \hline
        Foul & T2 & 0.127 \\
        Offside & T2 & 0.054 \\
        Direct free-kick & T2 & 0.040 \\
        Indirect free-kick & T2 & 0.049 \\
        \hline
        Ball out of play & T3 & 0.004 \\
        Throw-in & T3 & 0.014 \\
        Clearance & T3 & 0.006 \\
        Kick-off & T3 & 0.005 \\
        \hline
        \multicolumn{3}{l}{\textbf{Tier weights:} $w_{\text{T1}}=0.67$, $w_{\text{T2}}=0.27$, $w_{\text{T3}}=0.06$.} \\
        \hline
    \end{tabular}
\end{table}

\section{Derivation of Macro-Window Preference Alignment (MWPA)}
\label{app:mws_derivation}

This section derives Derivation of Macro-Window Preference Alignment (MWPA) from the standard RLHF formulation and shows that MWPA is a trajectory-level instantiation of DPO. Concretely, we keep the original DPO construction unchanged, but define the preference unit as a fixed-length macro-window rollout (a trajectory) rather than a single micro-step emission, which converts sparse streaming supervision into a stable event-level preference signal.

\subsection{Preliminaries: RLHF as KL-Regularized Reward Maximization}
Following \citet{christiano2017deep} and \citet{rafailov2023direct}, consider the KL-regularized RLHF objective:
\begin{equation}
\label{eq:rlhf}
\max_{\pi}\;
\mathbb{E}_{x \sim \mathcal{D}}
\left[
\mathbb{E}_{y \sim \pi(\cdot \mid x)} \left[r^*(y \mid x)\right]
-\beta\, D_{\mathrm{KL}}\!\left(\pi(\cdot \mid x)\,\|\,\pi_{\mathrm{ref}}(\cdot \mid x)\right)
\right],
\end{equation}
where $r^*$ is an (unknown) reward function reflecting human preference, $\pi_{\mathrm{ref}}$ is a fixed reference policy, and $\beta>0$ controls the deviation from the reference.

\subsection{Streaming to Trajectory-Level Conditioning and Responses}
In streaming commentary, the model processes a video stream incrementally and emits short text segments at each micro-step. For a macro window starting at step $k$ with horizon $L$, define:
\begin{itemize}
    \item $\mathcal{V}_{k:k+L-1} \triangleq (V_k, V_{k+1}, \ldots, V_{k+L-1})$: the video chunks in the window,
    \item $\mathbf{s}_{k-1}$: the shared streaming prefix state summarizing all history before step $k$ (e.g., the KV-cache / prefix representation induced by the interleaved past inputs under causal masking),
    \item $\mathcal{C}_k \triangleq (\mathbf{s}_{k-1}, \mathcal{V}_{k:k+L-1})$: the shared conditioning context for macro-window rollouts.
\end{itemize}

The macro-window response is a \textbf{trajectory} of emitted text segments:
\begin{equation}
\label{eq:trajectory_def}
Y_k \triangleq (W_k, W_{k+1}, \ldots, W_{k+L-1}),
\end{equation}
where each $W_t$ is a (possibly empty) token sequence emitted at micro-step $t$. Let $\mathrm{concat}(Y_k)$ denote the concatenation of all tokens emitted in the window (preserving causal temporal order and skipping silent or empty segments).

\paragraph{Autoregressive factorization.}
Because the policy is autoregressive, the trajectory likelihood factorizes over tokens. Writing $\mathrm{concat}(Y_k) = (y_1,\ldots,y_N)$, we have:
\begin{equation}
\label{eq:traj_factorization}
\log \pi_\theta(Y_k \mid \mathcal{C}_k)
\;\triangleq\;
\log \pi_\theta(\mathrm{concat}(Y_k) \mid \mathcal{C}_k)
=
\sum_{i=1}^{N} \log \pi_\theta\!\left(y_i \mid \mathcal{C}_k, y_{<i}\right).
\end{equation}
In implementation, although $\mathcal{C}_k$ includes the window video chunks for notational convenience, the model remains \emph{causal}: when predicting a token aligned to micro-step $t$, attention is restricted to visual evidence up to $V_t$ (via streaming state $\mathbf{s}_{t-1}$ and causal masking).

\subsection{Optimal Policy and Implicit Reward}
For a fixed $x$, the optimal solution to Eq.~\eqref{eq:rlhf} has the Boltzmann form \citep{rafailov2023direct}:
\begin{equation}
\label{eq:optimal_policy}
\pi^*(y \mid x) = \frac{1}{Z(x)} \pi_{\mathrm{ref}}(y \mid x)\exp\!\left(\frac{1}{\beta} r^*(y \mid x)\right),
\end{equation}
with partition function $Z(x)$. Rearranging yields the implicit reward representation:
\begin{equation}
\label{eq:implicit_reward}
r^*(y \mid x)
=
\beta \log \frac{\pi^*(y \mid x)}{\pi_{\mathrm{ref}}(y \mid x)} + \beta \log Z(x).
\end{equation}
In MWPA, we set $(x,y) \equiv (\mathcal{C}_k, Y_k)$, i.e., the ``response'' is a macro-window trajectory.

\subsection{From Pairwise Preferences to the DPO Objective (Trajectory Version)}
Assume the preference oracle follows a Bradley--Terry model:
\begin{equation}
\label{eq:bt}
p^*(Y_k^+ \succ Y_k^- \mid \mathcal{C}_k)
=
\sigma\!\left(r^*(Y_k^+ \mid \mathcal{C}_k) - r^*(Y_k^- \mid \mathcal{C}_k)\right),
\qquad
\sigma(z)=\frac{1}{1+e^{-z}}.
\end{equation}
Substituting Eq.~\eqref{eq:implicit_reward} into Eq.~\eqref{eq:bt} cancels the $\beta\log Z(\mathcal{C}_k)$ term and gives:
\begin{equation}
\label{eq:bt_logratio_pi_star}
p^*(Y_k^+ \succ Y_k^- \mid \mathcal{C}_k)
=
\sigma\!\left(
\beta\left[
\log \frac{\pi^*(Y_k^+ \mid \mathcal{C}_k)}{\pi_{\mathrm{ref}}(Y_k^+ \mid \mathcal{C}_k)}
-
\log \frac{\pi^*(Y_k^- \mid \mathcal{C}_k)}{\pi_{\mathrm{ref}}(Y_k^- \mid \mathcal{C}_k)}
\right]
\right).
\end{equation}

\paragraph{Key step (the missing bridge): model-implied preference and MLE.}
DPO does not require access to $\pi^*$. Instead, it defines a model-implied preference distribution $p_\theta$ by replacing $\pi^*$ in Eq.~\eqref{eq:bt_logratio_pi_star} with the learnable policy $\pi_\theta$, and fits it by maximum likelihood on observed preference pairs $(\mathcal{C}_k, Y_k^+, Y_k^-)$:
\begin{equation}
\label{eq:model_pref}
p_\theta(Y_k^+ \succ Y_k^- \mid \mathcal{C}_k)
=
\sigma\!\left(
\beta\left[
\log \frac{\pi_\theta(Y_k^+ \mid \mathcal{C}_k)}{\pi_{\mathrm{ref}}(Y_k^+ \mid \mathcal{C}_k)}
-
\log \frac{\pi_\theta(Y_k^- \mid \mathcal{C}_k)}{\pi_{\mathrm{ref}}(Y_k^- \mid \mathcal{C}_k)}
\right]
\right).
\end{equation}
Maximizing $\mathbb{E}_{\mathcal{D}}[\log p_\theta(\cdot)]$ yields the DPO loss.

\paragraph{Final MWPA loss.}
Define the trajectory-level log-probability gaps:
\begin{align}
\label{eq:delta_defs}
\Delta_\theta(\mathcal{C}_k; Y_k^+, Y_k^-)
&\triangleq
\log \pi_\theta(Y_k^+ \mid \mathcal{C}_k) - \log \pi_\theta(Y_k^- \mid \mathcal{C}_k), \\
\Delta_{\mathrm{ref}}(\mathcal{C}_k; Y_k^+, Y_k^-)
&\triangleq
\log \pi_{\mathrm{ref}}(Y_k^+ \mid \mathcal{C}_k) - \log \pi_{\mathrm{ref}}(Y_k^- \mid \mathcal{C}_k).
\end{align}
Then the Macro-Window Streaming DPO objective is:
\begin{equation}
\label{eq:mws_dpo_loss}
\mathcal{L}_{\mathrm{MWS}}(\theta)
=
-\mathbb{E}_{(\mathcal{C}_k,Y_k^+,Y_k^-)\sim \mathcal{D}}
\left[
\log \sigma\!\left(\beta\left(\Delta_\theta - \Delta_{\mathrm{ref}}\right)\right)
\right],
\end{equation}
where $\log \pi_\theta(Y_k \mid \mathcal{C}_k)$ is computed via the token-sum factorization in Eq.~\eqref{eq:traj_factorization}.

\subsection{Gradient Structure: Why Macro-Windows Reduce Variance}
Let $z \triangleq \beta(\Delta_\theta-\Delta_{\mathrm{ref}})$. Differentiating Eq.~\eqref{eq:mws_dpo_loss} gives:
\begin{equation}
\label{eq:grad_main}
\nabla_\theta \mathcal{L}_{\mathrm{MWS}}
=
-\beta\,\sigma(-z)\,
\left(
\nabla_\theta \log \pi_\theta(Y_k^+ \mid \mathcal{C}_k)
-
\nabla_\theta \log \pi_\theta(Y_k^- \mid \mathcal{C}_k)
\right).
\end{equation}
Using Eq.~\eqref{eq:traj_factorization}, the trajectory log-likelihood gradient decomposes into a sum over tokens within the macro window:
\begin{equation}
\label{eq:grad_token_sum}
\nabla_\theta \log \pi_\theta(Y_k \mid \mathcal{C}_k)
=
\sum_{i=1}^{N}
\nabla_\theta \log \pi_\theta\!\left(y_i \mid \mathcal{C}_k, y_{<i}\right).
\end{equation}

\paragraph{Implication for streaming granularity.}
In step-level streaming DPO ($L=1$), each preference comparison is supported by only a handful of emitted tokens, so the estimator in Eq.~\eqref{eq:grad_main} is dominated by high-variance micro emissions and often lacks sufficient semantic evidence for key events. In MWPA, Eq.~\eqref{eq:grad_token_sum} explicitly aggregates gradients across the entire macro window, turning sparse event-level preference information into a more stable update signal. Note that this remains window-level credit assignment: all tokens inside the rollout are updated coherently according to the trajectory preference, without requiring token-level preference labels.

\section{Limitations and Future Work}
\label{app:limitations}

A limitation of our work is that the current data and evaluation mainly focus on English commentary from SoccerNet-era broadcasts.
Extending SoccerNarrate to more recent seasons, more languages, and more diverse commentary styles remains an important direction for future work.
In addition, streaming commentary may still contain player or team name errors, especially under occlusion, fast transitions, or ambiguous visual evidence.
Future work could introduce stronger entity-level supervision, roster-aware decoding, and player identification modules to improve fine-grained factual reliability.
We also plan to extend the evaluation beyond key event grounding to cover tactical analysis, long-term narrative coherence, and broader commentary quality.
In addition, our model does not explicitly model color commentary from the second commentator, and a full-match human evaluation of overall commentary quality is left for future work.
Since SoccerNarrate is built from copyrighted broadcasts, we do not redistribute the raw video or audio, and derived annotations will be released only as permitted by SoccerNet and the rights holders.
Finally, automatic commentary may still contain incorrect event or player mentions, so practical deployment requires verification and human oversight.

\paragraph{Broader impacts.}
Automatic commentary could support sports coverage and make match information more accessible to audiences who benefit from textual descriptions.
However, incorrect event or player attribution, unsupported background facts, and uneven coverage across languages or commentary styles can mislead viewers or reproduce biases in the source material.
Generated text could also be reused in misleading automated media.
These risks motivate verification against the match context, clear disclosure that the commentary is machine-generated, and human editorial oversight before broadcasting.
We do not establish the suitability of SoccerNarrate for unsupervised live deployment.

\paragraph{Resources and availability.}
\label{app:resource_availability}
Our code and project resources are released at \url{https://github.com/MCG-NJU/SoccerNarrate}, with trained checkpoints and derived resources made available through the linked repositories.
Any release of transcripts and derived annotations remains subject to SoccerNet access conditions and the permissions of the broadcast rights holders; our work does not grant redistribution rights to the source video or audio or to third-party assets.

\end{document}